\documentclass[10pt,twocolumn,letterpaper]{article}

\usepackage{iccv}              
\usepackage{adjustbox}
\usepackage{fontawesome}
\usepackage{arydshln}
\usepackage{multirow}
\usepackage{marvosym, ifsym}
\usepackage{pifont}
\definecolor{iccvblue}{rgb}{0.21,0.49,0.74}
\usepackage[pagebackref,breaklinks,colorlinks,allcolors=iccvblue]{hyperref}

\def\paperID{3088} 
\def\confName{ICCV}
\def\confYear{2025}

\title{VRBench: A Benchmark for Multi-Step Reasoning in Long Narrative Videos}

\author{
Jiashuo Yu$^{1*}$ 
\quad Yue Wu$^{1*}$ 
\quad Meng Chu$^{1*}$ 
\quad Zhifei Ren$^{1*}$ 
\quad Zizheng Huang$^{2, 3, 1*}$ 
\quad Pei Chu$^{1*}$ \\
\quad Ruijie Zhang$^{1}$ 
\quad Yinan He$^{1}$ 
\quad Qirui Li$^{1}$ 
\quad Songze Li$^{1}$ 
\quad Zhenxiang Li$^{1}$ 
\quad Zhongying Tu$^{1}$  \\
\quad Conghui He$^{1}$ 
\quad Yu Qiao$^{1}$ 
\quad Yali Wang\textsuperscript{4, 1\Letter} 
\quad Yi Wang\textsuperscript{1, 3\Letter} 
\quad Limin Wang\textsuperscript{2, 1\Letter} \\
$^{1}$Shanghai Artificial Intelligence Laboratory  \quad $^{2}$Nanjing University \quad $^{3}$Shanghai Innovation Institute \quad\\ $^{4}$Shenzhen Institute of Advanced Technology, Chinese Academy of Sciences\\
{\tt \href{VRBench.github.io}{VRBench.github.io}}
}

\begin{document}
\twocolumn[{
            \renewcommand\twocolumn[1][]{#1}
            \maketitle
            \begin{center}
                \centering
                \includegraphics[width=0.95\textwidth]{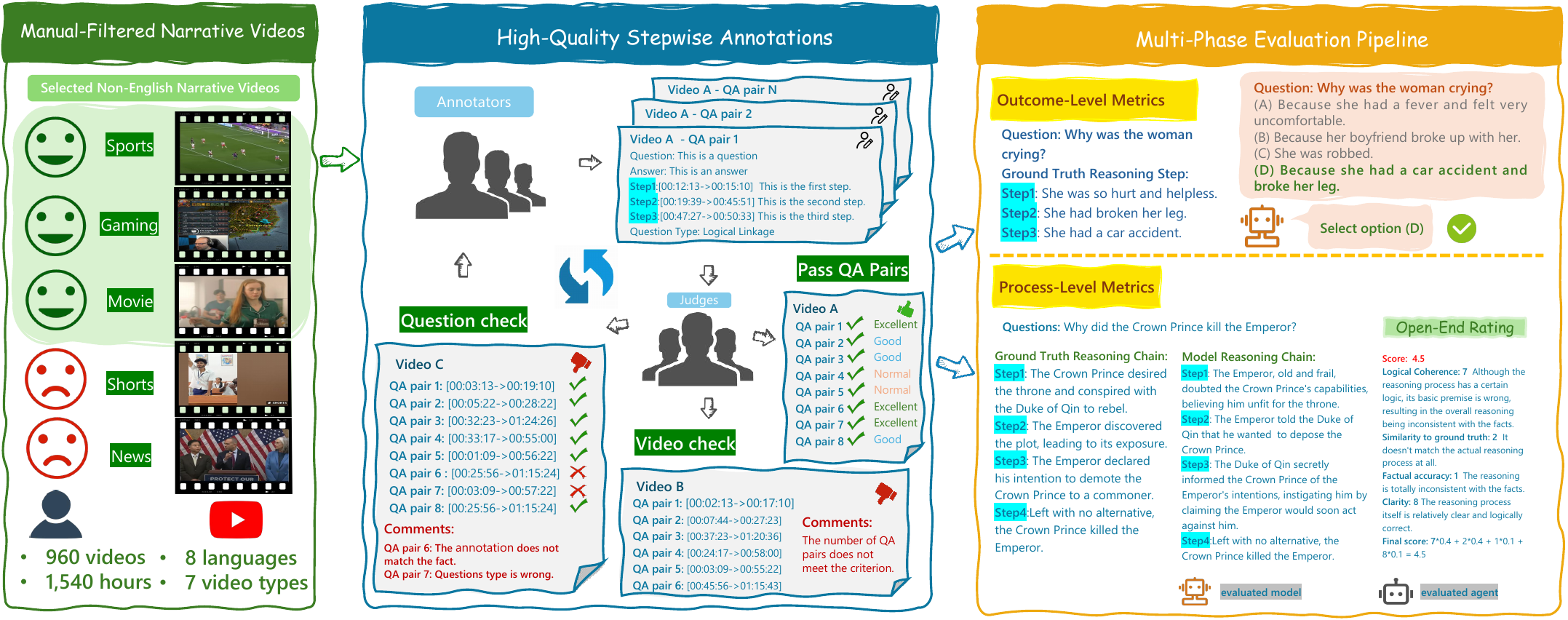}
                \captionof{figure} {
                \textbf{Overview of VRBench.} We present VRBench, a long narrative video benchmark for multi-step reasoning. VRBench includes 960 \textbf{manual-filtered narrative videos}, covering 8 languages and 7 video categories that are suitable for reasoning about temporal relations. We also provide~\textbf{high-quality stepwise annotations} for reasoning, which are labeled and reviewed by human experts. Each video incorporates 8-10 complex question-answer pairs, a multi-step reasoning chain, and fine-grained timestamps. To fully evaluate the capability of models in multi-step reasoning, we propose a~\textbf{multi-phase evaluation pipeline} that assesses model results both from the process and outcome levels. Our VRBench is the first video reasoning benchmark that supports both multi-step annotation and evaluation.
                }
                \label{fig:teaser}
            \end{center}
        }]

\begin{abstract}

\makeatletter{\renewcommand*{\@makefnmark}{}
\footnotetext{$^*$equal contributions. \textsuperscript{\Letter}corresponding authors.}\makeatother
}

We present \textbf{VRBench}, the first long narrative video benchmark crafted for evaluating large models' multi-step reasoning capabilities, addressing limitations in existing evaluations that overlook temporal reasoning and procedural validity. It comprises 960 long videos (with an average duration of 1.6 hours), along with 8,243 human-labeled multi-step question-answering pairs and 25,106 reasoning steps with timestamps. These videos are curated via a multi-stage filtering process including expert inter-rater reviewing to prioritize plot coherence. We develop a human-AI collaborative framework that generates coherent reasoning chains, each requiring multiple temporally grounded steps, spanning seven types (e.g., event attribution, implicit inference). VRBench designs a multi-phase evaluation pipeline that assesses models at both the outcome and process levels. Apart from the MCQs for the final results, we propose a progress-level LLM-guided scoring metric to evaluate the quality of the reasoning chain from multiple dimensions comprehensively. Through extensive evaluations of 12 LLMs and 19 VLMs on VRBench, we undertake a thorough analysis and provide valuable insights that advance the field of multi-step reasoning.
\end{abstract}

\section{Introduction}
\label{sec:intro}

The rapid evolution of vision language models (VLMs) has heightened the need for benchmarks that rigorously evaluate complex reasoning capabilities. While existing standards like GSM8K~\cite{gsm8k} focus on domain-specific knowledge in mathematics and science, they neglect a critical reasoning dimension: contextual analysis in visual narrative content. Real-world applications increasingly demand temporal reasoning across interconnected elements: tracking character dynamics in films, interpreting gameplay strategies, or understanding cause-and-effect chains in documentaries. This capability gap persists because current video benchmarks~\cite{videomme, lvbench, longvideobench, cinepile} primarily assess single-step perception rather than sustained reasoning processes.

We find three fundamental limitations in current evaluation paradigms: (1) an overemphasis on domain expertise rather than plot-driven reasoning, (2) the lack of temporally-grounded reasoning chains in annotations, and (3) Outcome-focused metrics that ignore procedural validity. To address these challenges, we present VRBench (as in Figure~\ref{fig:teaser}), the first benchmark specifically designed for multi-step reasoning in long-form narrative videos. Its construction involves the collection of narrative videos, a human-in-the-loop reasoning process method for annotation, and a multi-phase evaluation pipeline considering both procedure and outcome.

\begin{figure*}[ht]
  \centering
   \includegraphics[width=0.9\linewidth]{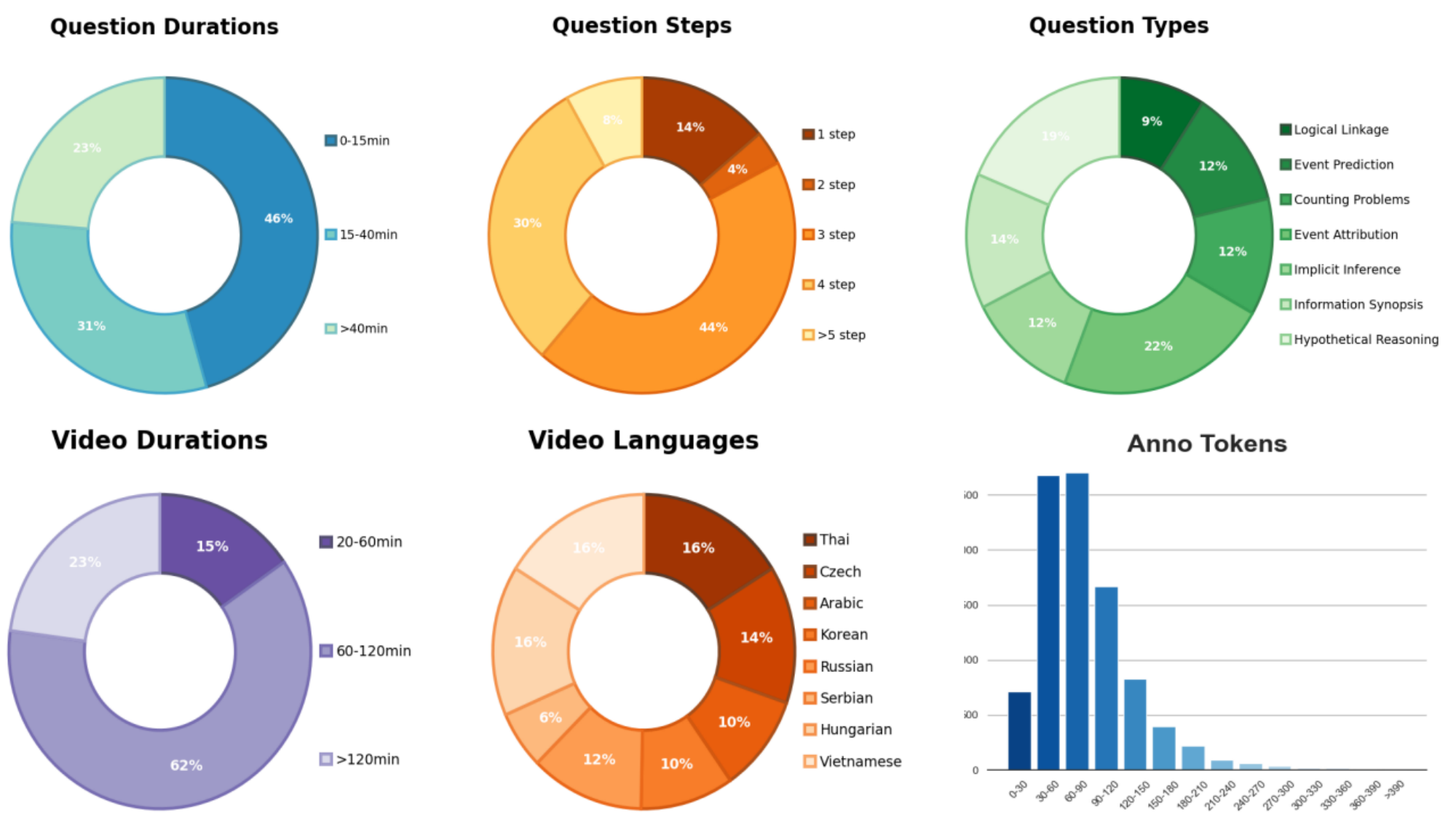}
   \caption{Statistics of VRBench. We provide the detailed distribution of videos and annotations of VRBench, including video languages and durations, steps, types, and temporal duration of questions, as well as the token numbers of answers and the reasoning process.}
    \label{fig:statistics}
\end{figure*}

In data construction, VRBench aggregates 960 meticulously selected videos (1.6h average duration) across seven narrative categories (e.g., movies, sports, travelogues), sourced through a multi-stage filtering process. Our curation pipeline combines automated retrieval with expert validation (inter-rater reliability $\rho$=0.82), prioritizing plot coherence over domain-specific knowledge. This contrasts with existing benchmarks like MMVU~\cite{mmvu} that emphasize disciplinary competence. To conduct multi-step reasoning chain annotation, we develop a human-AI collaborative approach, generating 8-10 QA pairs per video with explicit temporal grounding. Each question requires no less than 2 reasoning steps annotated with precise timestamps (Figure~\ref{fig:statistics}), validated through iterative expert review (95\% inter-annotator agreement). The taxonomy spans seven reasoning types from event prediction to implicit inference, ensuring comprehensive coverage of narrative analysis skills.

When evaluating models on VRBench, we propose a multi-phase evaluation pipeline that combines outcome verification with process analysis. Beyond conventional multiple-choice accuracy that evaluates the outcome-level performance, we further introduce the process-level metric LLM-guided scoring to evaluate the overall reasoning process quality. This dual approach reveals critical insights – for instance, while GPT-4o achieves 81.25\% outcome accuracy, its process rating scores drop to 56.11\%, exposing reasoning fragility. Evaluations of 31 state-of-the-art large models reveal proprietary VLMs with long-context support outperform text-only LLMs by 13.82\% absolute, emphasizing the importance of dense visual grounding. System-2 optimization strategies yield disproportionate gains in process metrics, and despite parameter parity, open-source VLMs lag behind proprietary counterparts by 12.30\%, suggesting architectural limitations beyond scale. Our ablation studies show strong alignment between human and LLM evaluations and quantify the impact of test-time scaling—doubling context windows improves QwQ-32B's~\cite{qwq} accuracy by 12.43\%.

VRBench sets a new standard for evaluating narrative reasoning, offering insights distinct from knowledge-centric benchmarks. By separating domain expertise from contextual analysis, our work aids in developing models capable of sustained, human-like understanding in real-world video content. The entire suite, including annotation protocols and evaluation tools, has been fully open-sourced to advance reasoning research.

\begin{figure}[t]
  \centering
   \includegraphics[width=0.95\linewidth]{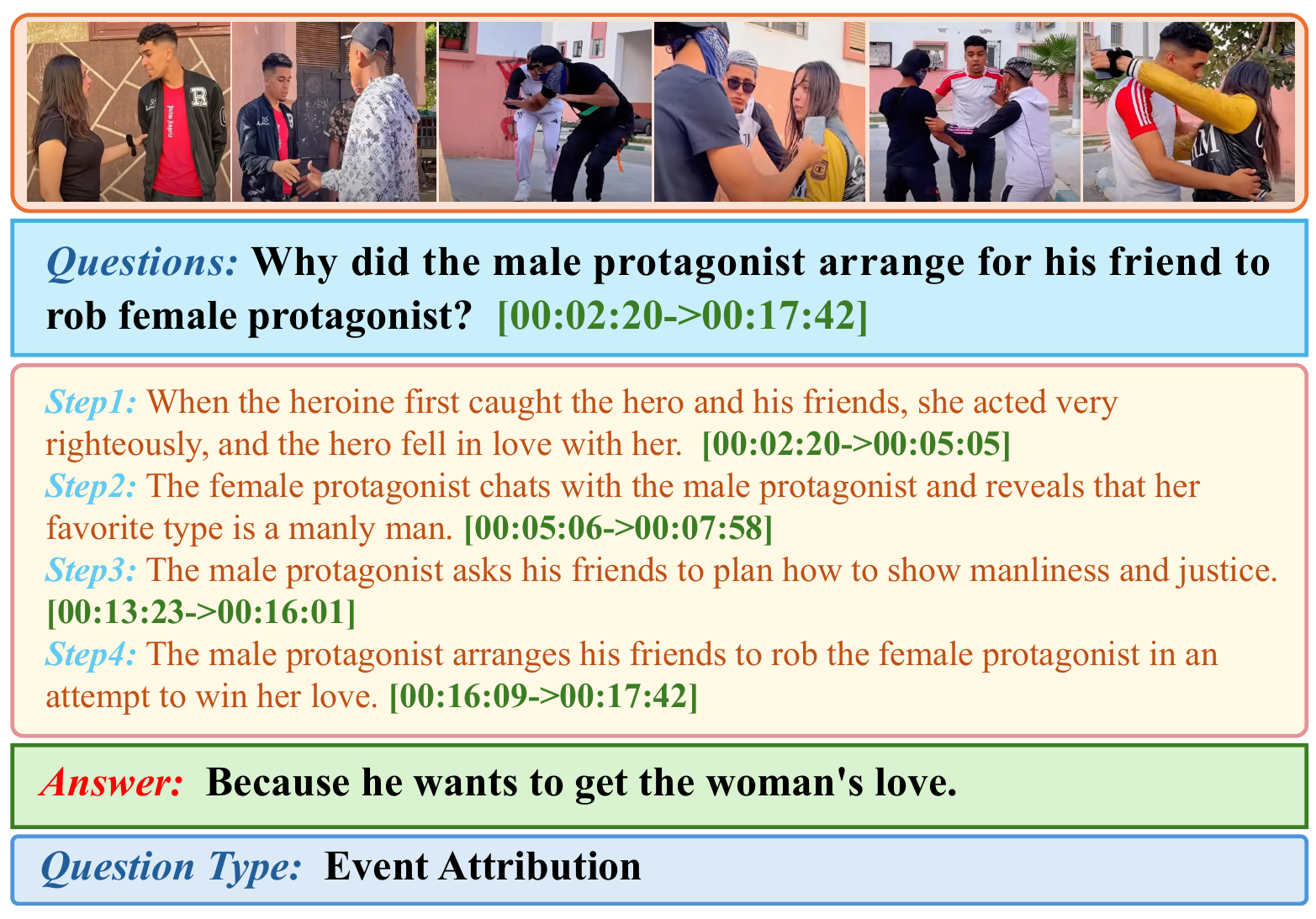}
   \caption{\textbf{An example of annotation in VRBench}. For each question, VRBench provides the question-answer pair, multi-step reasoning chain, question type, and the start-to-end timestamps of the entire question as well as each reasoning step.}
    \label{fig:example}
\end{figure}
\section{Related Work}
\label{sec:relatedworks}

\begin{table*}[htbp]
    \centering
\begin{adjustbox}{width=\linewidth,center}
\renewcommand{\arraystretch}{1.1}
\begin{tabular}{lccccccccccc}
\toprule  
{\textbf{Dataset}} & {\textbf{\#Size}} & {\textbf{\#QA pairs}} & {\textbf{\#Dur.(s)}} & {\textbf{QA types}} & {\textbf{Data Source}} & {\textbf{Anno Step}} & {\textbf{Eval Step}} & {\textbf{Eval Target}} & {\textbf{Anno Type}} & {\textbf{Clue}} & {\textbf{Multilingual}}  \\
\midrule
MMLU~\cite{mmlu} & 14,079 & 14,079 & / & MCQ & Multi-Disc & Single & Single & LLM & M & \faTimes & \faTimes \\
MMLU-Pro~\cite{mmlupro} & 12,032 & 12,032 & / & MCQ & Multi-Disc & Single & Single & LLM & A+M & \faTimes & \faTimes \\
LiveCodeBench~\cite{livecodebench} & 511 & 511 & / & MCQ & Code & Multi & Single & LLM & M & \faTimes & \faTimes \\
SciEval~\cite{scieval} & 15,901 & 15,901 & / & MCQ+Open & Science & Single & Single & LLM & M & \faTimes & \faTimes \\
GSM8k~\cite{gsm8k} & 8,792 & 8,792 & / & MCQ+Open & Math & Multi & Multi & LLM & M & \faTimes & \faTimes \\
C-Eval~\cite{ceval} & 12,342 & 12,342 & / & MCQ & Multi-Disc & Single & Single & LLM & M & \faTimes & \faTimes \\
\midrule
ScienceQA~\cite{scienceqa} & 10,332 & 21,208 & / & MCQ & Science & Multi & Single & VLM & M & \faTimes & \faTimes \\
VisScience~\cite{visscience} & 3,000 & 3,000 & / & MCQ+Open & Science & Single & Single & VLM & A+M & \faTimes & \faTimes \\
MMMU~\cite{mmmu} & 11,500 & 11,500 & / & MCQ+Open & Multi-Disc & Single & Single & VLM & M & \faTimes & \faTimes \\
MMMU-Pro~\cite{mmmupro} & 3,460 & 3,460 & / & MCQ & Multi-Disc & Single & Single & VLM & A+M & \faTimes & \faTimes \\
MathVista~\cite{mathvista} & 6,141 & 6,141 & / & MCQ+Open & Math & Multi & Single & VLM & M & \faTimes & \faTimes \\
MathVision~\cite{mathvision} & 3,040 & 3,040 & / & MCQ+Open & Math & Multi & Single & VLM & M & \faTimes & \faTimes \\
CharXiv~\cite{charxiv} & 2,323 & 11,615 & / & Open & Multi-Disc & Single & Single & VLM & M & \faTimes & \faTimes \\
OlympicArena~\cite{olympicarena} & 7,571 & 11,163 & / & MCQ+Open & Multi-Disc & Multi & Multi & VLM & M & \faTimes & \faTimes \\
\midrule
MVBench~\cite{mvbench} & 3,641 & 4,000 & 16.0 & MCQ & Open-Domain & Single & Single & VLM & A+M & \faTimes & \faTimes \\
EgoSchema~\cite{egoschema} & 5,063 & 5,063 & 180.0 & MCQ & Egocentric & Single & Single & VLM & A & \faTimes & \faTimes \\
LongVideoBench~\cite{longvideobench} & 3,763 & 6,678 & 473.0 & MCQ & Open-Domain & Single & Single & VLM & M & \faTimes & \faTimes \\
LVBench~\cite{lvbench} & 103 & 1,549 & 4,101 & MCQ & Open-Domain & Single & Single & VLM & M & \faTimes & \faTimes \\
CGBench~\cite{cgbench} & 1,219 & 12,129 & 1624.4 & MCQ+Open & Open-Domain & Multi & Multi & VLM & M & \faCheck & \faTimes \\
MLVU~\cite{mlvu} & 1,730 & 3,102 & 930 & MCQ & Narrative & Single & Single & VLM & M & \faTimes & \faTimes \\
VideoMME~\cite{videomme} & 900 & 2,700 & 1017.9 & MCQ & Open-Domain & Single & Single & VLM & M & \faTimes & \faCheck \\
Video-MMMU~\cite{videommmu} & 300 & 900 & 506.2 & MCQ & Multi-Disc & Single & Single & VLM & M & \faTimes & \faTimes \\
MMWorld~\cite{mmworld} & 1,910 & 6,627 & 108.0 & MCQ & Multi-Disc & Single & Single & VLM & A+M & \faTimes & \faTimes \\
MMVU~\cite{mmvu} & 1,529 & 3,000 & 51.4 & MCQ+Open & Multi-Disc & Multi & Single & VLM & M & \faTimes & \faTimes \\
\midrule
VRBench (Ours) & 960 & 8,243 & 5,796.0 & MCQ+Open & Narrative & Multi & Multi & LLM, VLM & M & \faCheck & \faCheck \\
\bottomrule
\end{tabular}
\end{adjustbox}
\caption{Comparison between VRBench and existing benchmarks. \#Size is the number of text/images/videos, \#Dur. means the average video duration, A and M indicate automatic and manual annotation type, respectively, and multi-disc denotes multi-disciplinary data source. Clue means clue-grounded annotation. Multilingual requires the number of data source languages to be greater than 2.
}
\label{tab:statistics}
\end{table*}

\paragraph{Reasoning Large Models.}
As the capabilities of LLMs continue to expand, frontier models have demonstrated significant potential in addressing high-order reasoning tasks. OpenAI pioneered this advancement with the release of o1~\cite{o1}, an LLM capable of sophisticated reasoning tasks. Subsequently, a series of proprietary~\cite{claude, geminithinking, o3-mini} and open-source~\cite{deepseekr1, internlm2, qwq} LLMs designed for advanced reasoning have emerged. Furthermore, researchers have shifted their focus towards developing VLMs possessing similar higher-order reasoning capabilities. Previously, some VLMs~\cite{internvl2.5, internvideo2.5, internvideo2, videochatflash, vila1.5, videollama3, longva, qwen2.5vl, aria} to a certain extent exhibited good reasoning proficiency due to their proficiency in tackling long-sequence context. More recently, a series of VLMs~\cite{llavao1, videocot, llamavo1, internvlmpo} trained on data incorporating higher-order Chains-of-Thought in RL-based algorithms~\cite{dpo, ppo, videochatr1} have been developed, enabling them to solve multidisciplinary, knowledge-intensive problems through a combination of fast and slow thinking processes.

\paragraph{Reasoning Benchmarks.}
With the rapid advancement of LLMs and VLMs in reasoning capability, numerous reasoning benchmarks have been developed in text and image-text modalities to facilitate comprehensive evaluation. For the textual domain, several knowledge-driven benchmarks~\cite{mmlu, mmlupro, ceval, scieval, theoremqa, sciknoweval} have been derived from exam and textbook data sources, and comprise expert-level questions that require multi-step reasoning. Meanwhile, some multimodal benchmarks also emerge to introduce multi-discipline tasks based on charts~\cite{chartqa, chartinsights, charxiv}, plots~\cite{plotqa}, exam~\cite{examsv, scemqa, mmmu, cmmmu, olympicarena}, or expert-level questions printed on the static images~\cite{visscience, scienceqa, mmsci, mmmupro}. It is noted that even in the image and text domains, benchmarks with process-level annotations~\cite{gsm8k, mrgsm8k} are relatively scarce. For video understanding, evaluation also gradually shifts from short video clip perception~\cite{mvbench, perceptiontest, veubench, tvqa, nextqa, autoevalvideo, tempcompass, videobench, tvbench} to long-form long video understanding~\cite{videomme, longvideobench, lvbench, egoschema, moviechat, movqa, movieqa, cinepile, etbench, mmbenchvideo} and single-step reasoning.~\cite{videoespresso, videomme, vript, cgbench, star, videovista}. More recently, some advances~\cite{mmbenchvideo, mmvu, mmworld} propose several new video reasoning tasks in multi-disciplinary scenarios like healthcare, engineering, and science. Different from previous works, VRBench is the first narrative video-based benchmark that is purely used for multi-step reasoning evaluation. Table~\ref{tab:statistics} further presents the detailed statistics of our dataset and distinguishes the difference between VRBench and existing text, image, and video reasoning benchmarks. 

\section{Benchmark}
\label{benchmark}

We present VRBench, a comprehensive multi-step reasoning benchmark consisting of a collection of long, multilingual, and narrative videos with corresponding question-answer pairs. Figure~\ref{fig:teaser} gives the overall pipeline of the benchmark construction, and we illustrate it as follows.

\subsection{Video Curation}
We collect long narrative videos from YouTube, yielding an initial pool of over 10,000 public videos. Considering plausible reasoning in steps requires videos with rich and coherent plots, we source long-form footage using a cherry-picked tag set and a comprehensive criterion for narration.

\paragraph{Plot-related Tags for Queries.} We employ a manually curated set of 7 semantic tags (e.g., Film \& Animation, Sport, Travel \& Events) to retrieve videos with strong narrative potential. This tag set is developed through iterative validation by domain experts, explicitly excluding non-narrative categories such as news broadcasts and lecture recordings. Our analysis revealed that these excluded categories exhibit limited visual-semantic dynamics (e.g., static camera angles in talk formats) and minimal event progression, rendering them ineffective for benchmarking temporal reasoning capabilities.

\paragraph{Criterion for Narration.} Our sourcing standards include video duration, its language source, and user ratings. The duration requirement (a minimum of 20 minutes and an average of 1.61 hours) ensures adequate temporal context for constructing reasoning chains. Our language diversity strategy intentionally excludes English and Chinese to counterbalance existing dataset biases, aligning with established findings that linguistic variety improves benchmark generalizability.

Upon the aforementioned two quantitative prerequisites, we organized a panel of 14 multilingual domain experts to evaluate candidate videos using a standardized 10-point scale based on plot coherence and richness. The scoring is guided by several instructions. Videos scoring below 7 are systematically excluded, resulting in a final curated set of 960 high-quality narratives. Detailed annotation protocols are documented in the supplementary materials.

\subsection{Stepwise Reasoning Annotations}
We label videos with reasoning steps via a two-stage human-in-the-loop framework. It first generates pseudo candidate QA pairs through automated pipelines, then employs expert-guided rewriting to curate high-quality annotations, ensuring rigorous adherence to benchmark specifications while maintaining multimodal reasoning fidelity. To further improve annotation quality, we also implement a comprehensive quality assurance.

\paragraph{Automatic Pipelines.} We first employ AutoShot~\cite{autoshot} to cut videos into several segments, and then use VideoChat2~\cite{mvbench} to caption them. For auditory content, we adopt whisper-large-v3~\cite{whisper} to obtain speech transcripts, and DeepL~\cite{deepl} to translate them into English. The video captions and translated subtitles are then put into GPT-4o~\cite{gpt4o} to generate 6 pseudo QA pairs with a multi-step reasoning process. To ensure the quality of reasoning, we specify 7 multi-step reasoning types for narrative videos as:
\begin{itemize}
    \item \textbf{Event prediction}: Forecast subsequent events in the video timeline.
    \item \textbf{Hypothetical reasoning}: Deduce plausible scenario outcomes from stated premises.
    \item \textbf{Event attribution}: Determine causal origins or underlying motivations of video events.
    \item \textbf{Implicit inference}: Extract unstated temporal, emotional, or relational context from visual cues.
    \item \textbf{Logical linkage}: Establish event-mediated connections between visual/narrative elements.
    \item \textbf{Information synopsis}: Condense critical information across multimodal inputs.
    \item \textbf{Counting problems}: Quantify state changes through arithmetic/combinatorial analysis.
\end{itemize}

\paragraph{Expert-guided Rewriting.} We recruit and train 67 graduate students to generate 8-10 high-quality QA pairs per video, providing raw video footage, translated subtitles, and GPT-generated pre-annotations that offer contextual hints without meeting final benchmark standards. Each QA pair must satisfy four rigorous criteria: (1) Non-synopsis questions require $\geq$2 timestamped reasoning steps (start/end times documented); (2) Temporal distribution constraints ($\leq$4 questions from 0-15min, $\geq$3 from 15-40min, $\geq1$ from 40-120min); (3) Coverage of $\geq$5 predefined reasoning taxonomies (from 7 categories) ensuring diversity; (4) Mandatory multimodal grounding where solutions demand both visual analysis (excluding subtitle-only answers) and explicit reasoning (beyond basic perception).

\paragraph{Comprehensive Quality Assurance.} To ensure annotations meet our stringent multimodal standards, we implement a rigorous verification protocol: 10 trained reviewers validate each annotation, with non-compliant entries returned to annotators for iterative revisions until fully compliant. Annotators and reviewers are exclusively recruited from top-tier universities to ensure academic rigor. We further enforce quality through dual safeguards: a systematic 5\% random sampling audit across both annotation and review stages, coupled with full documentation of protocols, annotation guidelines, and quality assessment criteria in supplementary materials.

\subsection{Multi-Phase Evaluation Pipeline}

We benchmark VLMs and LLMs through a comprehensive multi-phase evaluation pipeline, which compares predictions against ground-truth annotations both at the process and outcome level. For the outcome-level evaluation stage, we adopt a multiple-choice question (MCQ) format, where false options are generated by DeepSeek-V3~\cite{deepseekv3} using carefully-designed prompts based on human-annotated answers. For the process-level stage, we propose the open-ended rating to fully evaluate the model's multi-step reasoning capability. Specifically, we adopt an LLM to evaluate the overall quality of the whole reasoning process through four 0-10 scores: logical coherence (40\%), similarity to ground truth (40\%, excluded for event prediction and hypothetical reasoning tasks), factual accuracy (10\%), and clarity (10\%). DeepSeek-V3~\cite{deepseekv3} serves as the judge due to its optimal balance between cost and human alignment (Section~\ref{sec:experiments}). For the event prediction and hypothetical reasoning tasks, we argue that there is no ground-truth reasoning steps since there might be multiple possible predictions, hence, we remove the similarity score and compute the final score with an 8:1:1 weight. All metrics are normalized to 0-100 scales, and we report the rankings by computing average scores across all metrics.

\paragraph{LLM Evaluation Support.} For text-only LLMs, we convert videos to text inputs using Qwen2.5-72B-Instruct~\cite{qwen2.5} to synthesize video captions and subtitles into fine-grained summaries. The process involves: 1) Dividing inputs into 5-minute chunks for duplicate removal and abstraction; 2) Merging chunk-level abstracts into coherent summaries with detailed audio-visual descriptions. These summaries enable LLM evaluations while addressing context window limits for VLMs.
\begin{table*}[t]
    \centering
\begin{adjustbox}{width=\linewidth,center}
\renewcommand{\arraystretch}{1.05}
\begin{tabular}{lccccccccccc}
\toprule  
{\multirow{2}{*}{\textbf{Model}}} & {} {\multirow{2}{*}{\textbf{Overall}}} & \multicolumn{2}{c}{{\centering \textbf{Results by Metric}}}  & \multicolumn{7}{c}{{\centering \textbf{Results by Taxonomy}}}  \\
\cmidrule(lr){3-4}\cmidrule(lr){5-11}
{\textbf{}} & {\textbf{}} & {\textbf{MCQ-O}} & {\textbf{OE-P}} & {\textbf{EA}} & {\textbf{CP}} & {\textbf{HR}} & {\textbf{II}} & {\textbf{IS}} & {\textbf{EP}} & {\textbf{LL}} \\
\midrule
\textbf{LLMs} \\
\midrule
\textit{Proprietary Models} \\
GPT-4o~\cite{gpt4o} & 55.49 & 63.87 & 47.11 & 54.41 & 32.63 & 69.03 & 61.47 & 71.83 & 66.37 & 66.19 \\
Claude-3.7-Sonnet~\cite{claude} & 58.07 & 62.91 & 53.23 & 55.87 & 35.11 & 69.83 & 63.29 & 72.61 & 67.91 & 67.89 \\
o1-preview~\cite{o1} & 60.14 & \underline{68.47} & 51.81 & 56.97 & 35.87 & 70.51 & 64.81 & 73.13 & 68.27 & 69.41 \\
Gemini-2.0-Flash-Thinking~\cite{gemini} & \underline{60.97} & 67.53 & \underline{54.41} & \underline{58.61} & 38.11 & 72.09 & \underline{67.57} & \underline{74.19} & \underline{70.43} & \underline{71.13} \\
\hdashline
\textit{Open-Source Models} \\
QwQ-32B-preview~\cite{qwq-preview} & 35.90 & 27.51 & 44.29 & 34.31 & 34.41 & 46.29 & 39.67 & 27.97 & 42.39 & 40.61 \\
InternLM3-8B-Instruct~\cite{internlm2} & 47.81 & 50.31 & 45.31 & 45.47 & 34.87 & 60.33 & 51.77 & 55.47 & 56.97 & 56.81 \\
Qwen2.5-7B-Instruct~\cite{qwen2.5} & 48.29 & 52.61 & 43.97 & 46.93 & 35.17 & 61.83 & 51.57 & 54.87 & 56.63 & 57.03 \\
Llama3.3-70B-Instruct~\cite{llama3.3} & 49.84 & 52.59 & 47.09 & 47.63 & 38.07 & 63.87 & 54.03 & 54.83 & 59.47 & 56.97 \\
QwQ-32B~\cite{qwq} & 52.52 & 56.01 & 49.03 & 49.53 & 36.03 & 62.01 & 54.03 & 52.03 & 58.01 & 59.03 \\
Qwen2.5-72B-Instruct~\cite{qwen2.5} & 53.51 & 60.49 & 46.53 & 51.03 & 36.01 & 67.03 & 58.53 & 67.83 & 61.93 & 63.03 \\
DeepSeek-V3~\cite{deepseekv3} & 56.06 & 64.79 & 47.33 & 54.03 & 36.57 & 69.97 & 61.83 & 69.53 & 65.53 & 65.43 \\
DeepSeek-R1~\cite{deepseekr1} & 57.13 & 64.19 & 50.07 & 56.13 & \underline{38.91} & \underline{72.81} & 64.13 & 68.01 & 68.53 & 67.93 \\
\midrule
\textbf{VLMs} \\
\midrule
\textit{Proprietary Models} \\
Claude-3.7-Sonnet~\cite{claude} & 68.15 & 80.09 & 56.21 & 63.11 & 32.97 & 72.63 & 71.13 & 75.37 & 71.33 & 70.27 \\
GPT-4o~\cite{gpt4o} & 68.68 & 81.23 & 56.13 & 66.61 & 36.51 & 76.67 & 70.47 & 78.03 & 72.03 & 73.17 \\
Gemini-2.0-Pro~\cite{gemini} & \textbf{74.61} & \textbf{83.29} & \textbf{65.93} & \textbf{71.09} & \textbf{65.21} & \textbf{81.01} & \textbf{75.73} & 87.13 & \textbf{77.93} & \textbf{75.89} \\
\hdashline
\textit{Open-Source Models} \\
DeepSeek-VL2~\cite{deepseekvl2} & 31.50 & 33.27 & 29.73 & 25.93 & 22.41 & 35.73 & 31.73 & 30.01 & 33.29 & 29.57 \\
H2OVL Mississippi-2B~\cite{h2ovl} & 47.15 & 52.33 & 41.97 & 35.17 & 32.37 & 51.71 & 41.41 & 60.03 & 49.97 & 41.91 \\
Phi-3.5-Vision~\cite{phi} & 48.52 & 58.03 & 39.01 & 31.53 & 28.03 & 45.03 & 37.03 & 68.03 & 44.03 & 36.03 \\
LongVA-7B~\cite{longva} & 50.14 & 67.81 & 32.47 & 27.61 & 25.07 & 38.69 & 33.77 & 76.91 & 38.27 & 32.07 \\
InternVL2.5-8B~\cite{internvl2.5} & 50.41 & 69.31 & 31.51 & 26.63 & 26.31 & 37.99 & 31.21 & 85.97 & 37.37 & 30.69 \\
MiMo-VL-7B-RL~\cite{mimovl} & 50.48 & 63.39 & 37.57 & 47.31 & 34.59 & 63.83 & 53.23 & 74.61 & 60.33 & 56.97 \\
VideoChat-Flash-7B~\cite{videochatflash} & 50.82 & 72.01 & 29.63 & 24.69 & 21.41 & 39.37 & 29.17 & 79.37 & 34.87 & 28.07 \\
InternVideo2.5~\cite{internvideo2.5} & 51.94 & 75.63 & 28.25 & 23.81 & 24.09 & 34.51 & 27.07 & 84.57 & 33.61 & 26.81 \\
LongVA-7B-DPO~\cite{longva} & 52.36 & 67.91 & 36.81 & 30.73 & 27.17 & 45.09 & 38.23 & 79.39 & 43.67 & 37.03 \\
Qwen2-VL-7B~\cite{qwen2vl} & 54.08 & 72.01 & 36.15 & 30.49 & 26.61 & 46.71 & 33.69 & 85.27 & 43.31 & 35.29 \\
Aria~\cite{aria} & 54.55 & 72.97 & 36.13 & 30.19 & 29.23 & 44.23 & 34.63 & 86.23 & 44.99 & 34.57 \\
Qwen2.5-VL-7B~\cite{qwen2.5vl} & 56.52 & 69.61 & 43.43 & 37.07 & 33.17 & 54.27 & 42.27 & 83.91 & 50.17 & 42.87 \\
Keye-VL-8B-Preview~\cite{keyevl} & 60.44 & 64.41 & 56.47 & 62.53 & 40.37 & 73.61 & 67.99 & 70.13 & 71.83 & 71.03 \\
Qwen2.5-VL-72B~\cite{qwen2.5vl} & 61.71 & 66.85 & 56.57 & 51.87 & 46.13 & 67.13 & 54.13 & \textbf{90.04} & 63.67 & 60.77 \\
Kimi-VL-A3B-Thinking-2506~\cite{kimivl} & 61.82 & 61.67 & 61.97 & 64.53 & 47.91 & 71.37 & 69.57 & 74.47 & 71.23 & 71.47 \\
InternVL2.5-78B~\cite{internvl2.5} & 62.31 & 76.61 & 48.01 & 43.77 & 38.67 & 58.21 & 46.13 & 87.53 & 54.43 & 47.57 \\
\bottomrule
\end{tabular}
\end{adjustbox}
\caption{Evaluation results on VRBench across two evaluation metrics (outcome-level MCQ, process-level open-ended evaluation) and seven QA taxonomies (event attribution, counting problems, hypothetical reasoning, implicit inferences, information synopsis, event prediction, logical linkage). \textbf{Bold values} indicate the best results among all models, and \underline{underlined values} are best results of LLMs.}
\label{tab:results}
\end{table*}

\section{Experiments}
\label{sec:experiments}
We test a number of both open-source and proprietary VLMs (e.g. Qwen2-VL~\cite{qwen2vl}, InternVL2.5~\cite{internvl2.5}, InternVideo2.5~\cite{internvideo2.5}, GPT-4o~\cite{gpt4o}, Gemini-2.0-Pro~\cite{gemini}, etc.) and LLMs (DeepSeek-R1~\cite{deepseekr1}, QwQ~\cite{qwq}, OpenAI o1~\cite{o1}, Claude-3.7-Sonnet~\cite{claude}, etc.) with a two-stage evaluation on VRBench. The configuration of each evaluated model is detailed in the supplementary material. 

\noindent\textbf{Evaluation Protocols.}
Our protocol involves two phases. Given a question, 1) models produce multi-step reasoning using Chain-of-Thought~\cite{cot} prompts. Then 2) models choose final answers from multiple-choice options. For answer extraction, final answers are parsed from response endings to mitigate verbose option analysis, following~\cite{mvbench}. While reasoning process evaluations use judge LLMs with structured prompts and example-guided rating extraction. All evaluation prompts and implementation details are included in the supplementary materials.

\begin{figure}[t]
  \centering
   \includegraphics[width=\linewidth]{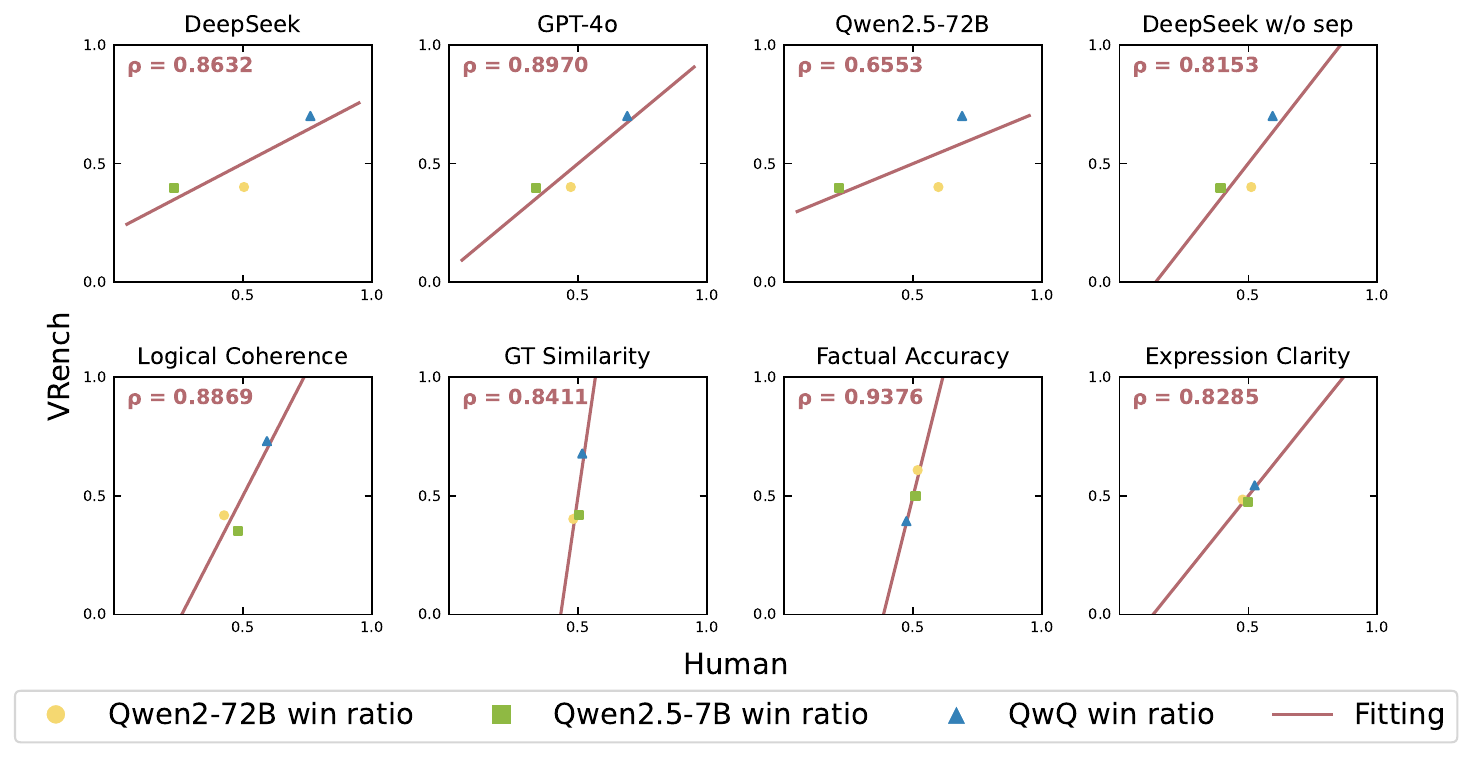}
   \caption{
    \textbf{Human preference alignment results.} For each plot, we show the win ratios of three different tested LLMs evaluated by human experts and VRBench. We then fit them with a straight line and quantify the correlation by calculating the Spearman correlation coefficient.
   }
    \label{fig:humanalign}
\end{figure}

\subsection{Main Results}

Table~\ref{tab:results} shows the CoT evaluation results of LLMs and VLMs on VRBench. Among LLMs, we found the proprietary model Gemini-2.0-Flash-Thinking~\cite{geminithinking} achieves the optimal performance while other reasoning models such as OpenAI o1-preview~\cite{o1} and DeepSeek-R1~\cite{deepseekr1} also demonstrate strong results. It is noted that o1-preview~\cite{o1} achieves the best outcome-level MCQ accuracy of 68.47\%. For the VLMs, Gemini-2.0-Pro~\cite{gemini} achieves 74.61\% overall accuracy, surpassing all components by at least 5.93\%. GPT-4o~\cite{gpt4o} and Claude-3.7-Sonnet~\cite{claude} also show competitive results, which demonstrate 68.68\% and 68.15\% overall performance, respectively. For open-source models, InternVL2.5-78B~\cite{internvl2.5}, emerges as the best non-proprietary VLMs with a 62.31\% overall accuracy.

We then delve into the specific results of each metric and taxonomy. For results in each evaluation stage, we find that most of LLMs and VLMs are capable of achieving high MCQ accuracy, yet compared to LLMs, VLMs struggle to demonstrate their reasoning steps and exhibit lower reasoning ratings. Through results across taxonomy, we found that most of the large models are not proficient in counting problems since they require fine-grained visual perception, in contrast with other tasks. Since the perception of LLMs fully depends on the video summary and some VLMs only support a small number of frame inputs, they are unable to correctly perceive the original frames containing the target elements, resulting in the MCQ accuracy close to random guessing and low reasoning process ratings.

\subsection{Ablations and Analysis}

We then further discuss the observations from the evaluation results and highlight some insights on obtaining higher multi-step reasoning performance.

\noindent\textbf{Does the evaluation of VRBench align with Human Preference?}
In contrast to MCQs that compute accuracy via a deterministic method, the correctness of process-level open-ended evaluation highly depends on the judgment of LLM's performance, which may include hallucinations that could influence the evaluation reliability. To this end, we select a subset of 30 videos with 300 questions to perform process-level human evaluation, aiming to quantify the correlation between human and LLM assessment results.

Specifically, the human annotators are asked to give process-level open-ended ratings following the same requirement for the judging LLM, i.e., four separated ratings are needed for each question to evaluate the logical coherence, similarity to ground-truth, factual accuracy, and expression clarity. We then compute the win ratio of pairwise comparison for each tested model following ~\cite{vbench}, where a model scores 1 if its rating is higher than its current opponent, 0 if lower, and 0.5 for a tie. Finally, we calculate the average win ratio of each model and the Spearman correlation coefficients ($\rho$) of the win ratios evaluated by LLMs and human annotators.

We first investigate the human preference alignment of several LLMs. We select GPT-4o~\cite{gpt4o}, DeepSeek-V3~\cite{deepseekv3}, Qwen2.5-7B~\cite{qwen2.5}, and Qwen2.5-72B~\cite{qwen2.5} as evaluation models and give them the same rating prompt for a fair comparison. As shown in Figure~\ref{fig:humanalign}, Qwen2.5 shows low human preference alignment compared with DeepSeek-V3 and GPT-4o, which both have correlation coefficients greater than 0.8. Though the human alignment of GPT-4o is slightly better than DeepSeek-V3, we observe the evaluation cost of GPT-4o is approximately 10 times that of DeepSeek, which imposes a significant burden on the benchmark users. Hence, we adopt DeepSeek-V3 as the judging model for its optimal trade-off between performance and cost. 

We also probe the evaluation correctness of each open-ended sub-metric. Following the same procedure, we compute the win ratio and correlation coefficient of DeepSeek-V3 for each sub-metric, and the results are shown in Figure~\ref{fig:humanalign} (b-d). Results show the coefficients in all metrics are higher than 0.8, indicating a strong positive monotonic relationship between LLM and human evaluation results. Another alternative is to instruct the LLM to compute the final score based on the given weights of each sub-metric, which is denoted as \textit{DeepSeek w/o sep} in Figure~\ref{fig:humanalign}. Results show that adding this additional computing process diminishes the robustness of evaluation LLM, thus in practice, we ask the model to output all sub-metric rating separately and automatically compute the final score.

\noindent\textbf{The role of System-2 thinking.} o1-like LLMs, also denoted as system-2 models, have emerged with superior reasoning abilities in multi-disciplinary scenarios such as science and math. We further explore these models' performance on the narrative video benchmark. We both assess the role of open-source and proprietary o1-models, as well as the gap compared with their previous system-1 version. For proprietaries, we compare OpenAI o1-preview~\cite{o1} with GPT-4o~\cite{gpt4o}, and results show that the system-2 version o1 outperforms GPT-4o with 4.30\% overall accuracy. For open-source model DeepSeek-V3~\cite{deepseekv3} and R1~\cite{deepl}, we observe that DeepSeek-R1 achieves higher scores both on multiple-choice questions and open-ended evaluations. It is noted that compared with Qwen2.5-72B-Instruct~\cite{qwen2}, QwQ-32B~\cite{qwq} gets higher reasoning ratings yet lower outcome-level MCQ results. This shows that though some system-2 models are more proficient in multi-step thinking, their ability to converge lengthy thinking processes into correct answers still needs improvement. Recently, some o1-like training strategies have also been utilized by some VLMs, such as LongVA-7B-DPO~\cite{longva}, a model that utilizes direct preference optimization~\cite{dpo} for long video understanding. It achieves an overall performance of 52.36\% and surpasses its vanilla version, suggesting that training VLMs with different optimization methods is feasible and that developing system-2 VLM models is crucial for tackling multi-step reasoning questions in narrative videos.

\noindent\textbf{Impact of Model Size.} It is commonly believed that models with large scales are often accompanied by better perception and reasoning capabilities. To verify such claims, we conduct experiments on the models with different scales, such as Qwen2.5-7B~\cite{qwen2.5} and Qwen2.5-72B, Qwen2.5-VL~\cite{qwen2.5vl} with 7B and 72B parameters, and InternVL2.5-8B~\cite{internvl2.5} and InternVL2.5-78B. As shown in Table~\ref{tab:results}, models with more parameters obtain higher overall accuracy (48.29\% vs 53.51\% for Qwen2.5, 56.52\% vs 61.71\% for Qwen2.5-VL, and 50.41\% vs 62.31\% for InternVL2.5), and the performance gap on process-level metrics is more apparent. This indicates that large-scale models are more likely to achieve advanced reasoning capabilities. However, we also noticed that models specifically trained on the reasoning corpus like QwQ-32B~\cite{qwq}, can achieve comparable performance compared with its 72B Qwen2.5 counterpart (52.52\% vs. 53.51\% in overall accuracy) with a smaller parameter amount. This suggests that training small-sized models with various preference optimization strategies and reasoning-centric corpora might improve their overall reasoning capability.

\noindent\textbf{Long Context Helps.} We observe a substantial performance gap between models with and without long context support. As shown in Table~\ref{tab:results}, models with long frame input tend to achieve higher overall accuracy. For example, Gemini-2.0-Pro~\cite{gemini} is capable of inference with 0.5 fps and supports large number frame inputs when answering questions of videos over 1 hour, and it achieves the optimal performance compared with all other fixed-length models. For the open-source models, Qwen2.5-VL-7B~\cite{qwen2.5vl} and Kimi-VL-A3B-Thinking-2506~\cite{kimivl} with 256 frame input achieve 61.71\% and 61.82\% overall accuracy, which is 11.30\% and 11.41\% higher than the 8B InternVL-2.5~\cite{internvl2.5} model with 64 frames input. Since the average length of VRBench is 1.61 hours, the necessity of long context support is becoming more pronounced, so that models can perceive more plot elements that help the analysis of long video content.

\noindent\textbf{LLMs vs. VLMs.} Since tested LLMs are only provided with the automatically-generated video summary, yet VLMs are capable of directly perceiving the raw visual content, it is unfair to directly compare the numerical results of LLMs and VLMs. However, there are still some noteworthy phenomena that can be discussed. First, VLMs with coarse-grained visual input (e.g., 4 frames for H2OVL Mississippi-2B~\cite{h2ovl} and DeepSeek-VL2~\cite{deepseekvl2}) perform significantly worse than several system-2 LLMs like DeepSeek-R1~\cite{deepseekr1} and Claude-3.7-Sonnet~\cite{claude}. This indicates that randomly selecting a small number of frames is not enough to tackle narrative-based reasoning tasks that require long-range perception. Furthermore, VLMs with fine-grained visual perception, such as Gemini-2.0-Pro with a 0.5 fps perform better than top-tier LLMs on both process-level and outcome-level metrics, showing the inevitability of detailed visual content input for tackling questions in VRBench.

\subsection{Test-Time Scaling Exploration}

\begin{figure}[t]
  \centering
   \includegraphics[width=0.95\linewidth]{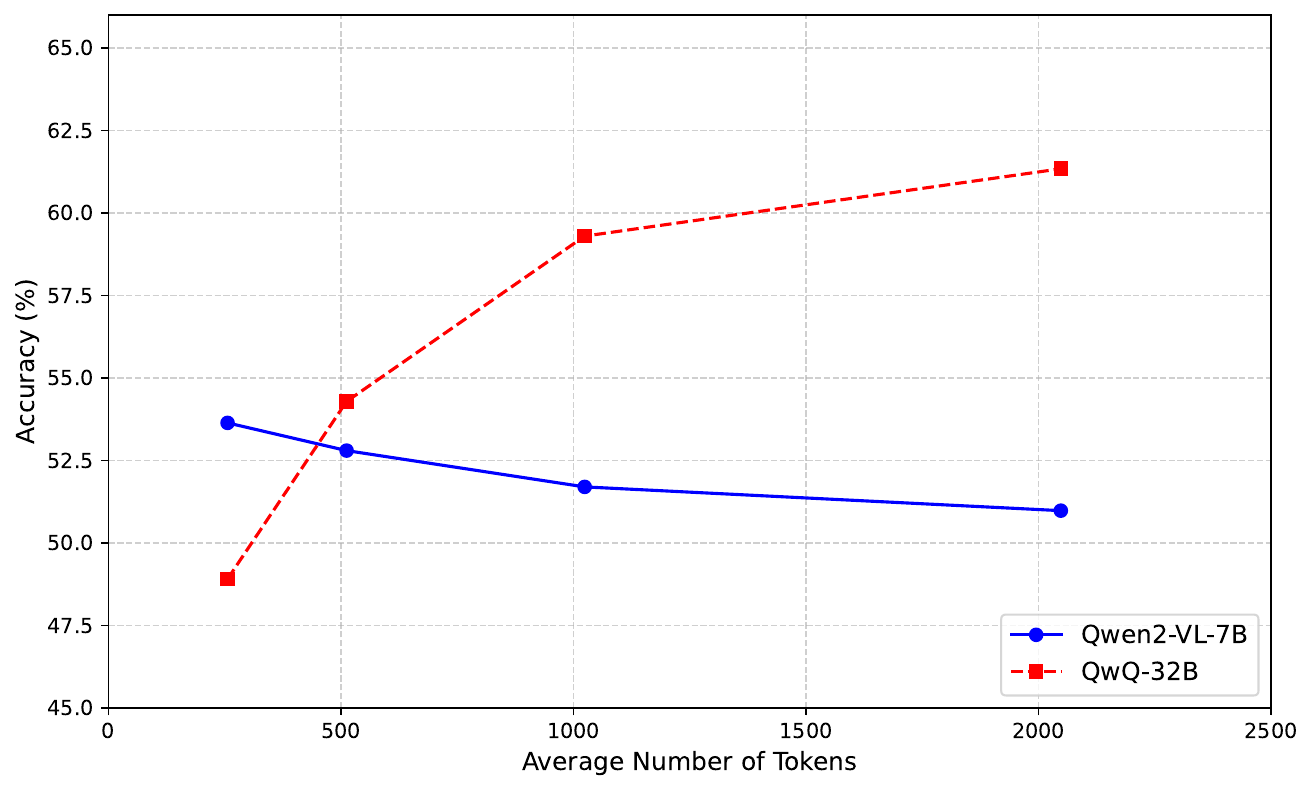}
   \caption{
    Test-Time Scaling Results. We report the average accuracy of outcome-level MCQ and process-level open-ended ratings.
   }
    \label{fig:tts}
\end{figure}

We here probe models' performance when scaling test-time compute cost. Since we tend to prompt the models with a Chain-of-Thought template when tackling complex reasoning tasks, it is observed that system-2 models are capable of dramatically expanding the inference budget to improve their performance. To this end, we set a series of token limitations on these models and require the models to think with different instructions. Specifically, we select QwQ-32B~\cite{qwq} as the experimental LLM and Qwen2-VL-7B~\cite{qwen2} as the testing VLM on a subset with 300 videos and 2,403 questions. For each model, several parallel experiments are conducted with increasing maximum token limits from 256 to 2048. We instruct the models to output the reasoning process as much as possible for large token limitations, and let them think concisely when the token limit is low. Since we aim to investigate the quality of models' initial reasoning process and their capability to generate the right answers through thinking, we report the overall scores that both evaluate the outcome accuracy and process quality. Results are shown in Figure~\ref{fig:tts}, where QwQ shows a remarkable rating boost from 48.91\% to 61.34\% when setting a large token limit. On the contrary, the small-sized system-1 model Qwen2-VL-7B performs even worse when using a large token number limitation and instructing the model to output more thinking process. The model tends to output more ambiguous outputs that lead to the wrong answer. This observation suggests that models with large parameter scales and system-2 capabilities benefit from the test-time scaling strategy. It also provides the insight that developing more test-time scaling approaches that guide the model to generate long reasoning traces could be a feasible and promising way to unlock more potential capabilities of system-2 models, thereby benefiting complex tasks that require multi-step reasoning.

\section{Conclusion}

This paper introduces VRBench, a comprehensive long-narrative video benchmark for evaluating multi-step reasoning. Through manually filtered narrative videos, high-quality stepwise annotations, and a multi-phase evaluation pipeline, VRBench distinguishes itself from other existing reasoning benchmarks and shows the robust capability to evaluate LLMs and VLMs both from the process and outcome perspectives. By assessing and analyzing 31 frontier large models, we thoroughly demonstrate the detailed performance of current reasoning models across various reasoning questions and provide valuable insights towards constructing more advanced multi-step reasoning models.

\section{Acknowledgement.} This work is supported by the National Key R$\&$D Program of China (No. 2022ZD0160101) and Jiangsu Frontier Technology Research and Development Program (No. BF2024076),
{
    \small
    \bibliographystyle{ieeenat_fullname}
    \bibliography{main}
}
\newpage
\clearpage
\maketitlesupplementary

\renewcommand\thesection{\Alph{section}}
\renewcommand\thefigure{A\arabic{figure}}
\renewcommand\thetable{A\arabic{table}}

\section{Question Exemplar}

Due to page limitation, we only provide one qualitative demo of our annotation QA pairs in the main manuscript, hence we complement more examples regarding all question types in this section in Figure A6-A12. 

\begin{figure}[htbp]
  \centering
   \includegraphics[width=0.95\linewidth]{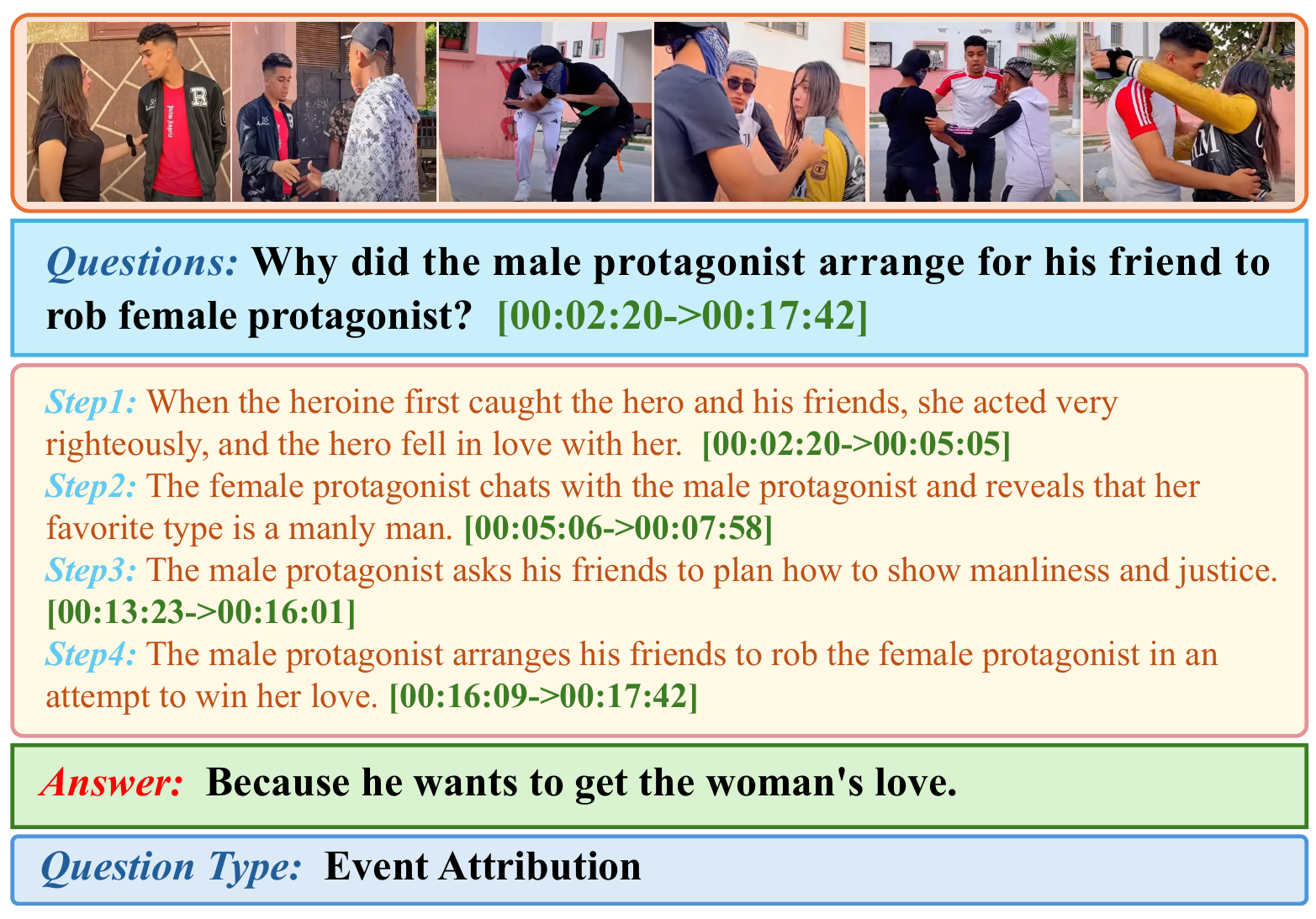}
   \caption{\textbf{Event attribution}: Determine causal origins or underlying motivations of video events.}
    \label{fig:event_attribution}
\end{figure}
\begin{figure}[htbp]
  \centering
   \includegraphics[width=0.95\linewidth]{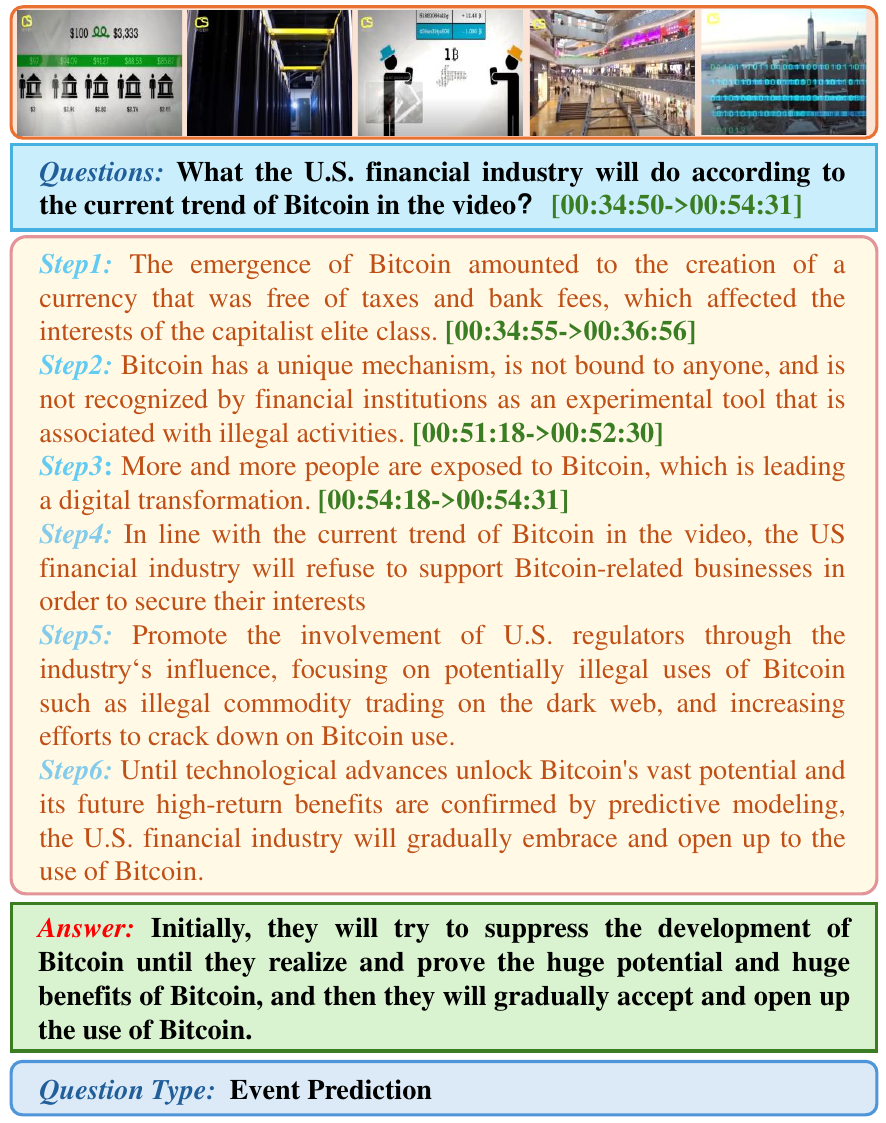}
   \caption{\textbf{Event prediction}: Forecast subsequent events in the video timeline.}
    \label{fig:event_prediction}
\end{figure}
\begin{figure}[htbp]
  \centering
   \includegraphics[width=0.95\linewidth]{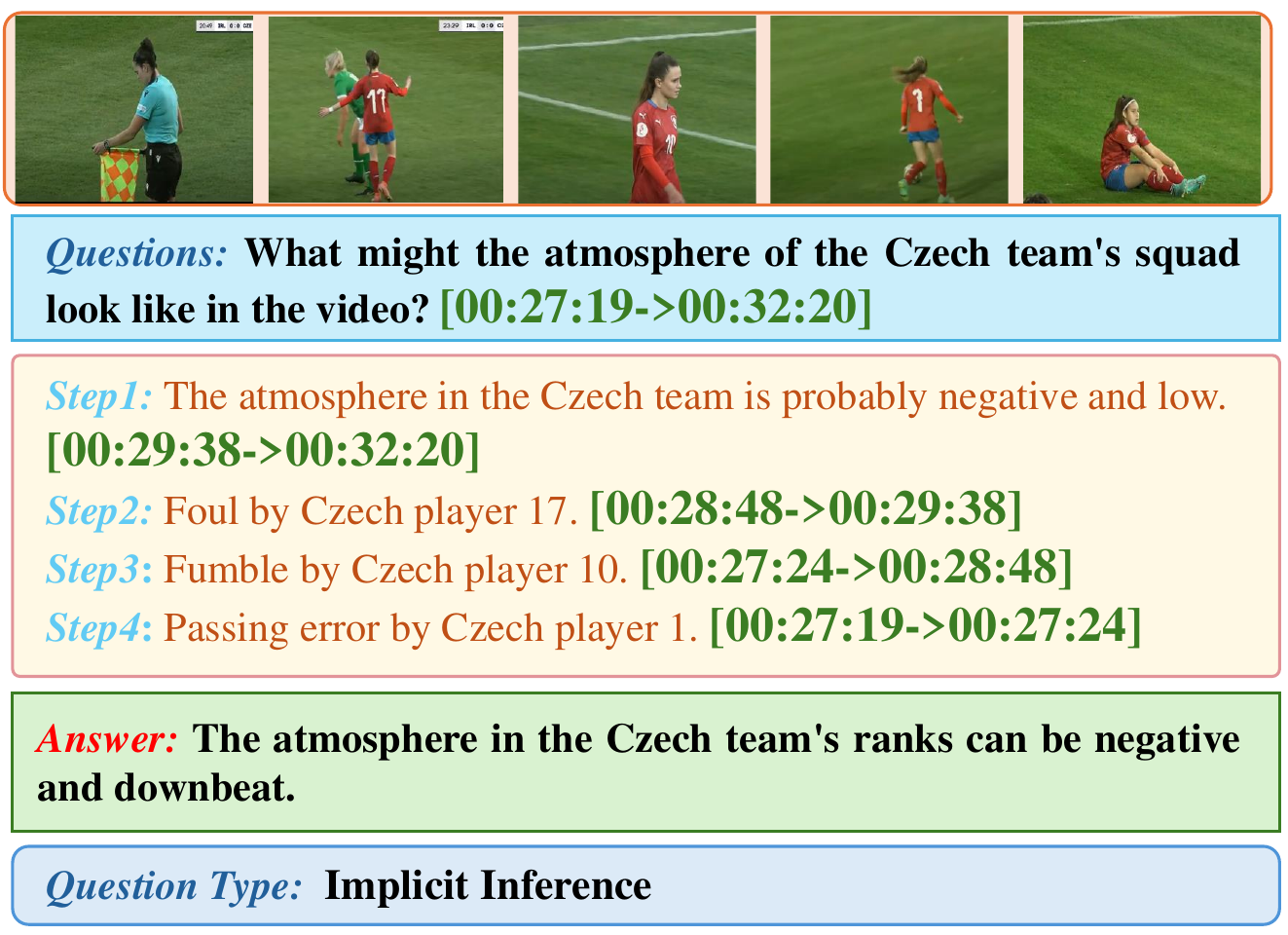}
   \caption{\textbf{Implicit inference}: Extract unstated temporal, emotional, or relational context from visual cues.}
    \label{fig:implicit_inference}
\end{figure}
\begin{figure}[htbp]
  \centering
   \includegraphics[width=0.95\linewidth]{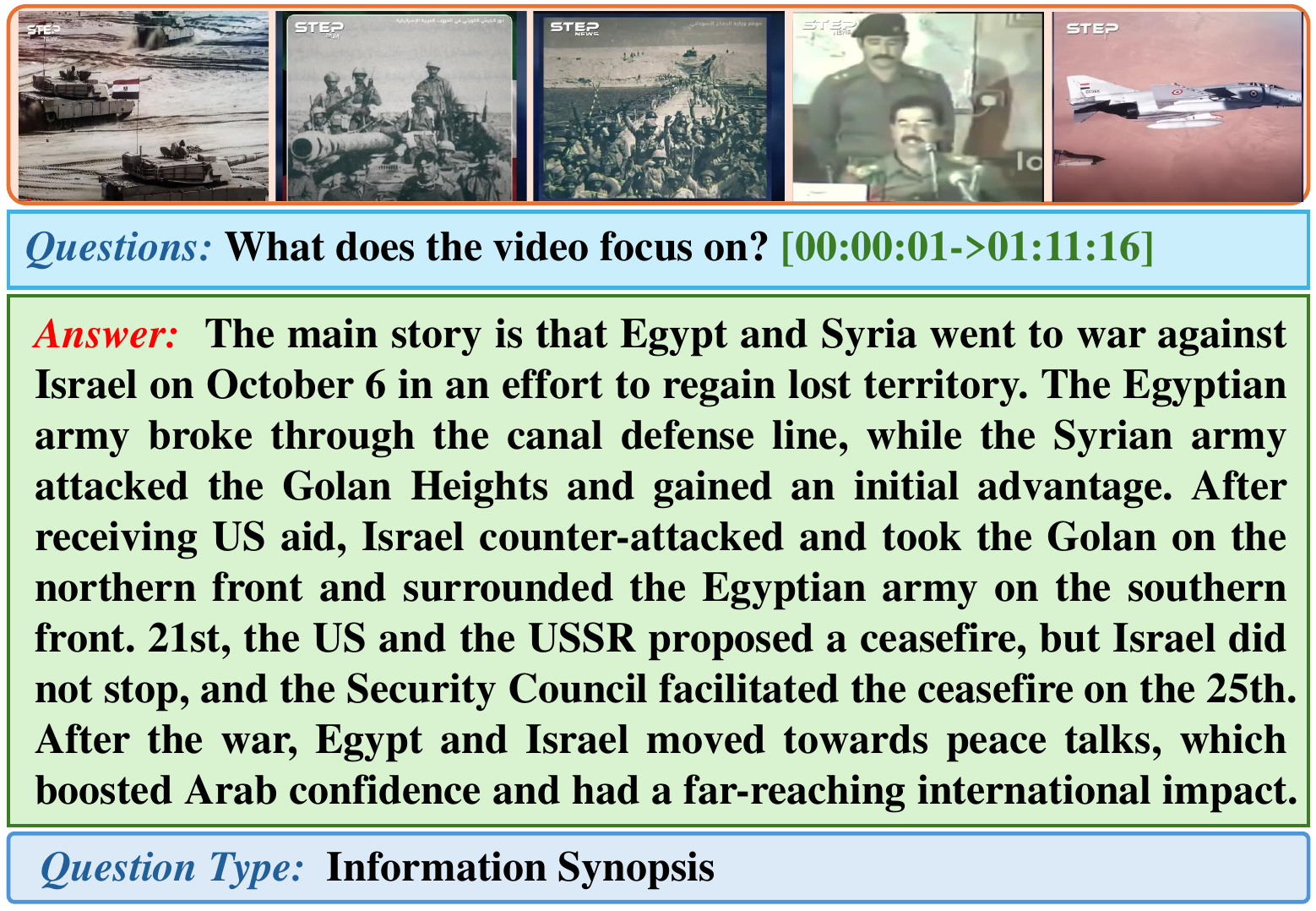}
   \caption{\textbf{Information synopsis}: Condense critical information across multimodal inputs.}
    \label{fig:information_synopsis}
\end{figure}
\begin{figure}[htbp]
  \centering
   \includegraphics[width=0.95\linewidth]{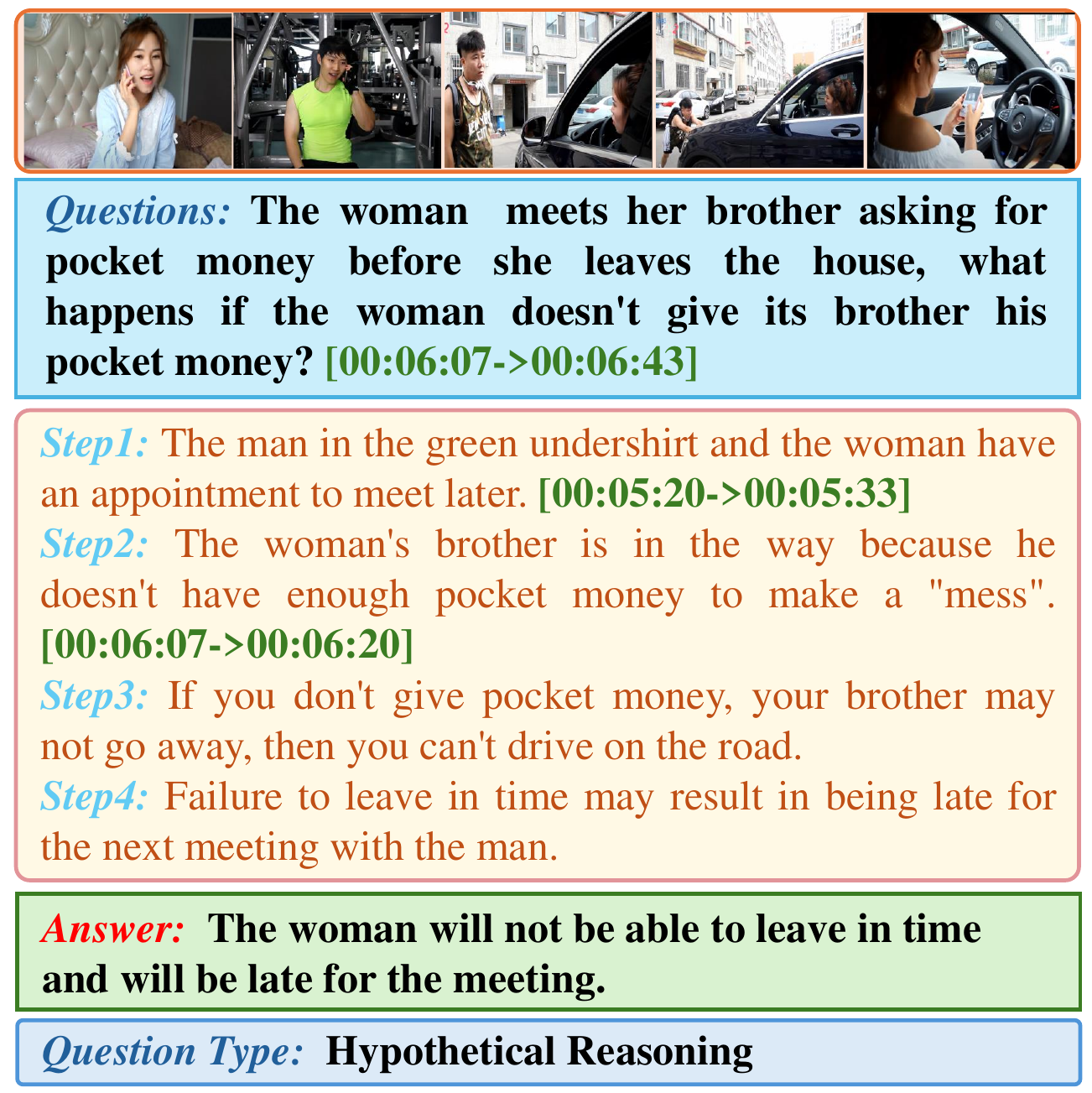}
   \caption{\textbf{Hypothetical reasoning}: Deduce plausible scenario outcomes from stated premises.}
    \label{fig:hypothetical_reasoning}
\end{figure}
\begin{figure}[htbp]
  \centering
   \includegraphics[width=0.95\linewidth]{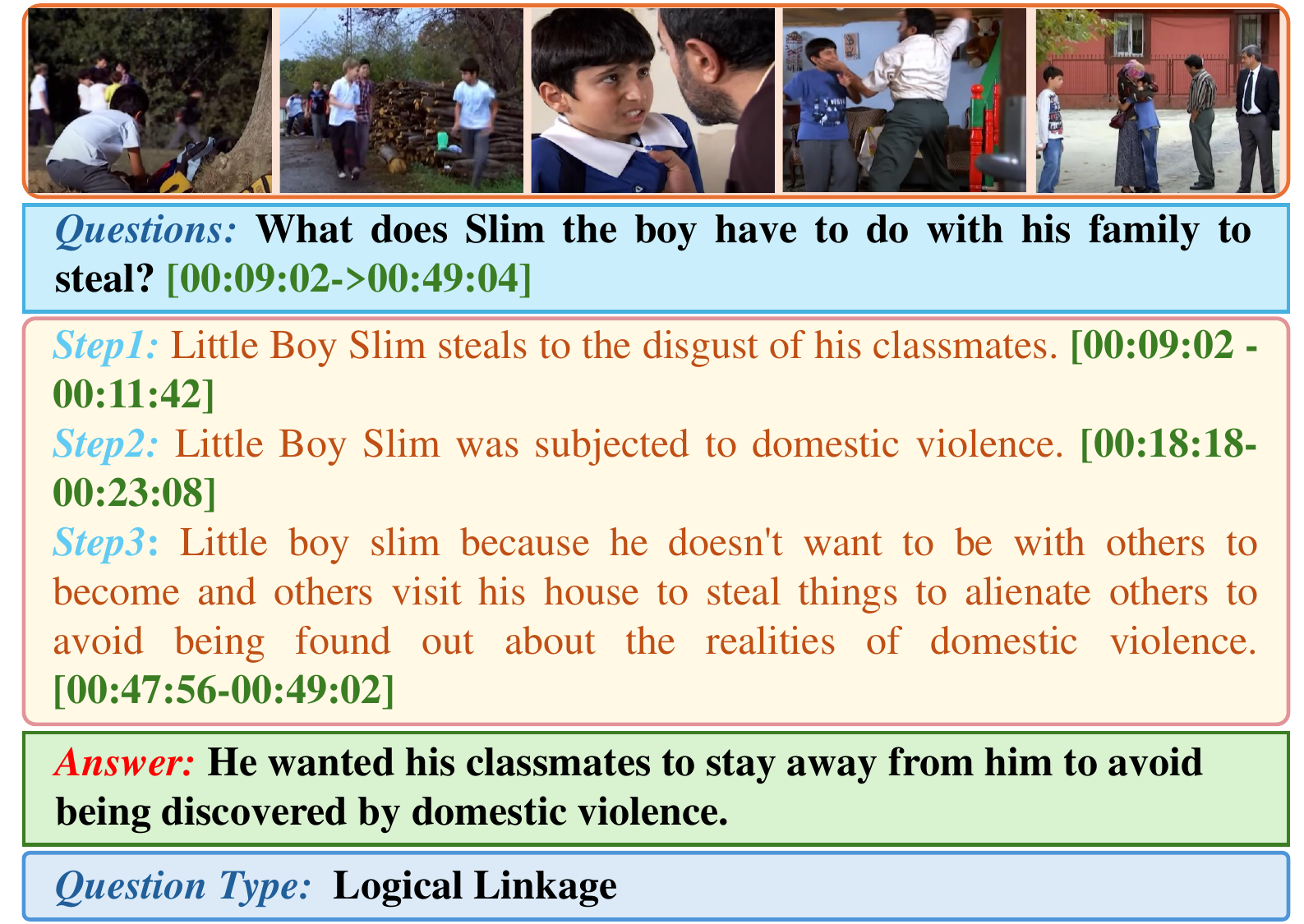}
   \caption{\textbf{Logical linkage}: Establish event-mediated connections between visual/narrative elements.}
    \label{fig:logical_linkage}
\end{figure}
\begin{figure}[htbp]
  \centering
   \includegraphics[width=0.95\linewidth]{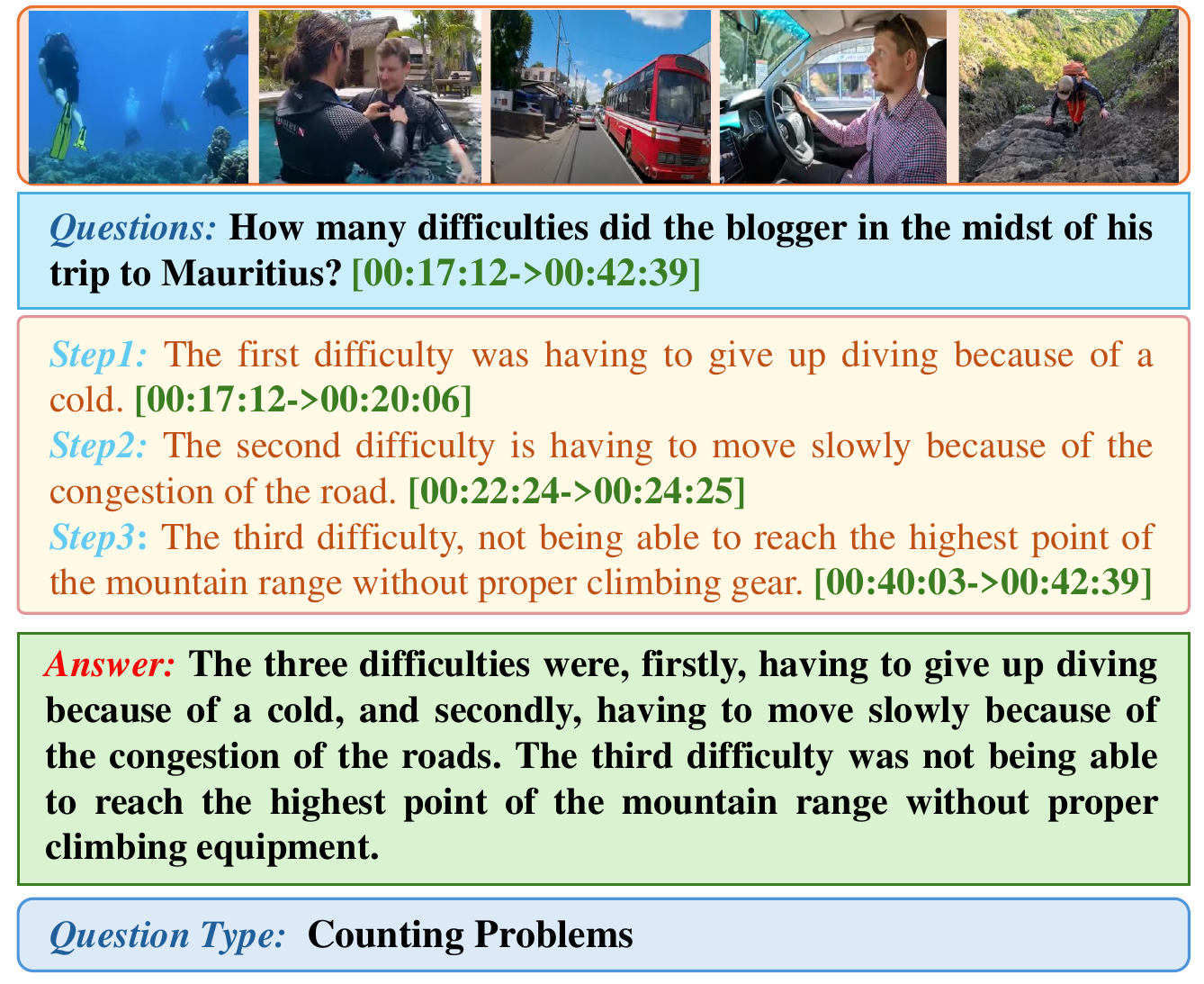}
   \caption{\textbf{Counting problems}: Quantify state changes through arithmetic/combinatorial analysis.}
    \label{fig:counting_problems}
\end{figure}

\section{Manual Filtering Details}

In this part, we provide additional details about the video manual-filtering progress, including the complete annotation guideline and the interface screenshot.

\subsection{Annotation Guideline}

\subsubsection{Confidentiality Statement}

The content of this document involves trade secrets or other secrets, and is only for internal members and authorized units or individuals to review, please receive this document to keep it in a safe place, without the consent of any third party shall not be disclosed to this document.

\subsubsection{Task Overview}

Given a video with corresponding translated subtitles, the annotators need to:
\begin{itemize}
    \item Quickly browse long videos and subtitles and determine their video category.
    \item Depending on the video category, rate the long video from 1 to 10 according to the category-specific rating guideline.
    \item Give an overall description of the video and the reason for the rating.
\end{itemize}

\noindent\textbf{Overall Rating Rules.} The overall range of scoring is from 1-10, with higher scores representing more suitable for constructing multi-step reasoning questions. A score of 7-10 means this video is very suitable for constructing multi-step reasoning pairs, a score of 4-6 represents it is moderately suitable, and a score of 1-3 represents a complete lack of suitability.

\noindent\textbf{Score distribution requirements.} There should not be a situation where one/some of the categories are all above or below 7 points. Separate score judgments based on the current video category according to the category-specific rating criteria.

\noindent\textbf{ASR Transcripts Quality Check.} Please note that annotators must use the video content in conjunction with the ASR transcript to determine whether reasoning is appropriate or not. If the video contents are static (e.g., speech, PPT presentation, two people sitting and chatting) and there is a lot of reasonable content in the speech, it should be labeled as unsuitable for multi-step reasoning annotation.

\begin{figure}[t]
  \centering
   \includegraphics[width=0.95\linewidth]{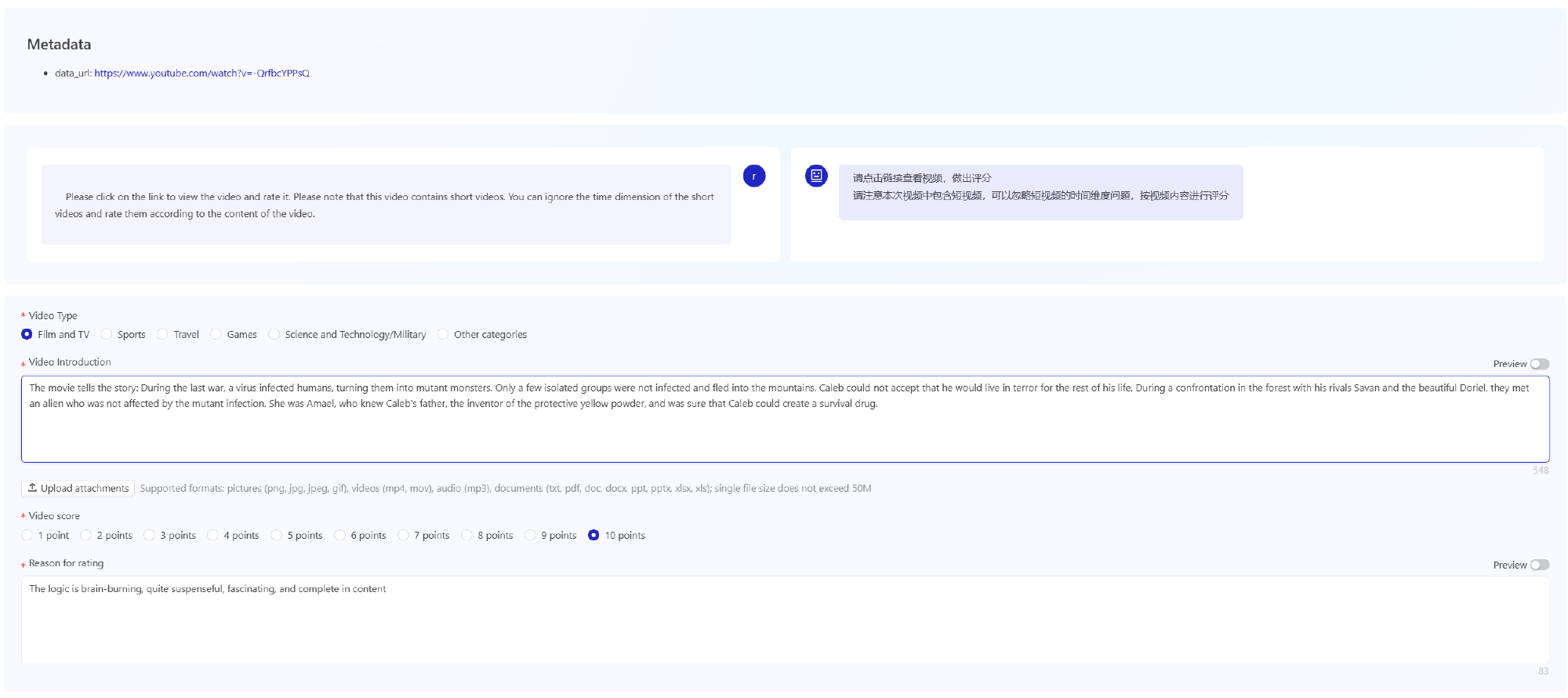}
   \caption{Interface of manual video filtering.}
    \label{fig:filter_interface}
\end{figure}
\begin{figure}[t]
  \centering
   \includegraphics[width=0.95\linewidth]{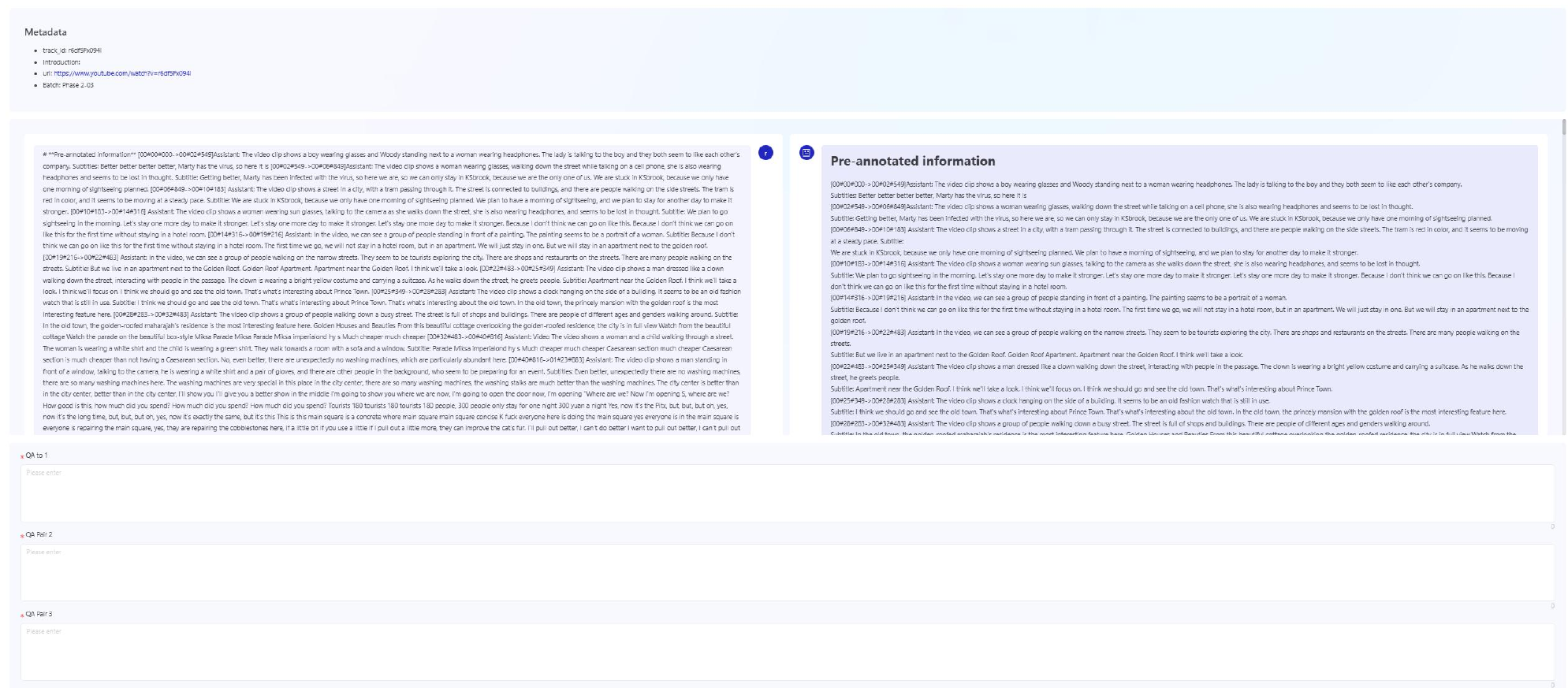}
   \caption{Interface of manual video labeling.}
    \label{fig:annotation_interface}
\end{figure}

\subsubsection{Rating Standard}

\noindent\textbf{Movies and Animation.}
\begin{itemize}
    \item Plot coherence: if it is a movie/TV series/animation with a coherent and complete plot, it is suitable for labeling, if it is an episode/spliced fragmented video, it is not suitable.
    \item Content richness: if it's a long documentary/landscape movie/music MV, it's not suitable for multi-step reasoning. If it contains a rich plot, it is suitable for labeling
    \item Transcripts correctness: the translated subtitle should be clear and readable, there may be some inaccurate transcripts, but the overall content must be understandable.
\end{itemize}

\noindent\textbf{Sports videos}

\begin{itemize}
    \item Sports category: If it is a video with small visual change content such as golf, racing, riding, etc., it is not suitable for multi-step reasoning annotation. If it is a video containing rich technical and tactical content and formation change, such as basketball, soccer, volleyball, etc., it is suitable for constructing reasoning tasks.
    \item Episode coherence: Suitable for labeling if it is a coherent and complete sports video, not suitable if it is a collection/spliced fragmented video.
    \item Transcripts correctness: The translated subtitle should be clear and readable, there may be some inaccurate transcripts, but the overall content must be understandable. In addition, the subtitles need to contain some explanation of the sports content to facilitate the understanding of the sports content and the construction of QA questions
    \item First/third person: Videos in first person is much less likely to construct multi-step reasoning problems, while third/rebroadcast view allows to gather details of the overall lineup and makes it easier to construct reasoning problems
\end{itemize}

\noindent\textbf{Travel \& Vlogs}

\begin{itemize}
    \item Episode Coherence: must be a complete travel video that can be spread out over multiple days, but not a compilation of multiple vlogs of multiple people in the same location, nor a compilation of vlogs of one person in multiple locations
    \item Content richness: Travel content needs to have a clear plot twist, in addition to some relevance in terms of long time granularity to construct long time questions. If it is a running travel record or contains only a record of scenery, it is not suitable for multi-step reasoning annotations.
    \item Transcripts correctness: the translated subtitle should be clear and readable, there may be some inaccurate transcripts, but the overall content must be understandable.
\end{itemize}

\noindent\textbf{Gaming Videos}

\begin{itemize}
    \item Game Type: Games with rich mechanics are suitable for constructing reasoning problems, such as MoBA games, RPG games, etc. Puzzle games are not suitable for constructing reasoning tasks.
    \item Content coherence: A full game match is good for constructing reasoning questions while the highlight collection is not suitable.
    \item First/Third View. For role-playing games, first view videos accompanied by an anchor's explanation are suitable for constructing deduction missions, but for multiplayer online games, the third-person view is more suitable for constructing qa questions, and the first-person view is not suitable.
    \item Game length: Full-length games are better suited for constructing reasoning questions, long videos spliced together from multiple short-length games are not suitable, such as Fortnite/PUBG anchor recordings with no obvious contextual connection.
    \item Transcripts correctness: the translated subtitle should be clear and readable, there may be some inaccurate transcripts, but the overall content must be understandable.
\end{itemize}

\noindent\textbf{Science and technology/military videos}

\begin{itemize}
    \item Plot coherence: Must be a complete scientific/military video, try to choose scientific and military explanatory videos, not news and reports.
    \item Content richness: In conjunction with subtitle content, there needs to be a clear reasoning element, in addition to some relevance at long time granularity for constructing long time questions
    \item Transcripts correctness: the translated subtitle should be clear and readable, there may be some inaccurate transcripts, but the overall content must be understandable.
\end{itemize}

\noindent\textbf{Other videos} For other categories of video that may appear, such as news and interviews, the rating should not be higher than 3 in principle. In practice, the annotator may also make subjective judgments on videos in some rarely occurring category, and may give high scores to videos with high reasoning content as appropriate, but the score should not exceed 8.

\subsubsection{Quality Assessment}

The quality assessors will conduct a 10:1 sampling ratio to check whether the video scoring meets the QA standards and the overall score distribution requirements.

\subsection{Annotation Interface}

The annotation interface is illustrated in Figure~\ref{fig:filter_interface}

\section{Human Annotation Details}

In this section, we complement more human annotation details, including the complete annotation guideline and the interface screenshot.

\subsection{Annotation Guideline}

\subsubsection{Confidentiality Statement}
The content of this document involves trade secrets or other secrets, and is only for internal members and authorized units or individuals to review, please receive this document to keep it in a safe place, without the consent of any third party shall not be disclosed to this document.

\subsubsection{Task Overview}
Given a video with subtitles and 6 pseudo pre-annotations, generate 8-10 question-answer pairs that require multi-step reasoning based on the video content and provide the detailed reasoning process with both question and stepwise timestamps.

\subsubsection{Annotation Standard}
Please follow the process below for labeling:

\noindent\textbf{Preliminaries.} We provide 6 pseudo pre-annotations generated by GPT-4o~\cite{gpt4o}. Please evaluate their authenticity and generate new questions based on these pre-annotations.

\noindent\textbf{Annotation Requirements.}
\begin{itemize}
    \item The questions posed must be reasoning-based, not perception-based. Answers should rely on logical reasoning and require an understanding of the overall plot development, rather than simple scene/object comprehension.
    \item The answers to the question are expected to be relatively fixed.
    \item Except for information synopsis type, answers require multi-step reasoning. The number of reasoning steps is equal to the number of reasoning processes - 1 (For example, if the reasoning process only contains the vanilla question and answers, the reasoning step number is counted as 1.). No more than 2 single-step reasoning questions are allowed per video.
    \item Questions and answers must be derived solely from the video content or a combination of video and subtitles, not from subtitles alone or common life knowledge.
    \item Different question types: Each video's 8-10 QA pairs must include 5 out of the 7 provided question types. No more than 1 event summary-type question is allowed.
    \item Different time granularities: Design questions that can be answered based on diverse video lengths, divided into [0-15 minutes, 15-40 minutes, 40 minutes-2 hours]. 
    \begin{itemize}
        \item For the 0-15 minutes questions, reasoning questions can be single-step, as multi-step questions may not be feasible. No more than 4 questions should be from the 0-15 minute interval.
        \item Videos shorter than 20 minutes can be marked as "unanswerable" and will be approved without review.
        \item Videos shorter than 50 minutes do not require questions from the 40 minutes-2 hours; only questions from the 0-15 minutes and 15-40 minutes intervals are needed.
    \end{itemize}
\end{itemize}

\noindent\textbf{Multi-Step Reasoning Expressions.} We provide the logical expressions of each multi-step question type to facilitate the understanding of multi-step reasoning.
\begin{enumerate}
    \item \textbf{Event Prediction}: Predict the next event based on event A that has already occurred in the video.
    \begin{itemize}
        \item Multi-step process 1: A → B → C → D.
        \item Multi-step process 2: B → A → C → D if A is attributed by B.
    \end{itemize}
    
    \item \textbf{Hypothetical Reasoning}: Given a hypothetical premise A, infer the corresponding development.
    \begin{itemize}
        \item Multi-step process 1: A → B → C → D.
        \item Multi-step process 2: B → A → C → D if A is attributed by B.
    \end{itemize}
    
    \item \textbf{Event Attribution}: Analyze the cause or purpose of event D in the video.
    \begin{itemize}
        \item Multi-step process: D → C → B → A.
    \end{itemize}
    
    \item \textbf{Implicit Reasoning}: Infer the feelings/emotions of a specific character D, relationships between characters, or the situation of event development at the current point in time.
    \begin{itemize}
        \item Multi-step process: D → C → B → A.
    \end{itemize}
    
    \item \textbf{Logical Links}: Analyze the correlation between two elements A and B in the video and explain their logical relationship.
    \begin{itemize}
        \item Multi-step process: A → C → D → B.
    \end{itemize}
    
    \item \textbf{Information synopsis}: Pose a summary question based on the video content and attempt to answer it (note: the question should not simply summarize the entire video but should be a synopsis question based on the video, requiring only single-step reasoning).
    
    \item \textbf{Counting Problems}: Infer the transformation of element A under multiple conditions, possibly involving arithmetic or counting components such as numbers, dates, or specific points in time.
    \begin{itemize}
        \item Multi-step process: A $\circ$ B $\circ$ C $\circ$ D, where $\circ$ represents any logical/mathematical operation.
    \end{itemize}
\end{enumerate}

\noindent\textbf{Annotation Format.} The annotation format should be as follows:
\begin{verbatim}
Timestamp: [xx:xx:xx->xx:xx:xx]
Question: xxxx
Answer: xxxx
Reasoning Process:
1. [xx:xx:xx->xx:xx:xx]
2. [xx:xx:xx->xx:xx:xx]
3. [xx:xx:xx->xx:xx:xx]
Reasoning Type: xxxx
\end{verbatim}

\subsubsection{Quality Assessment}
We adopt a full-scale quality assessment strategy, and the unqualified annotations should be modified until they meet the qualified criteria. We detail the quality assessment process in Section~\ref{appen:qa}.

\begin{figure}[t]
  \centering
   \includegraphics[width=0.95\linewidth]{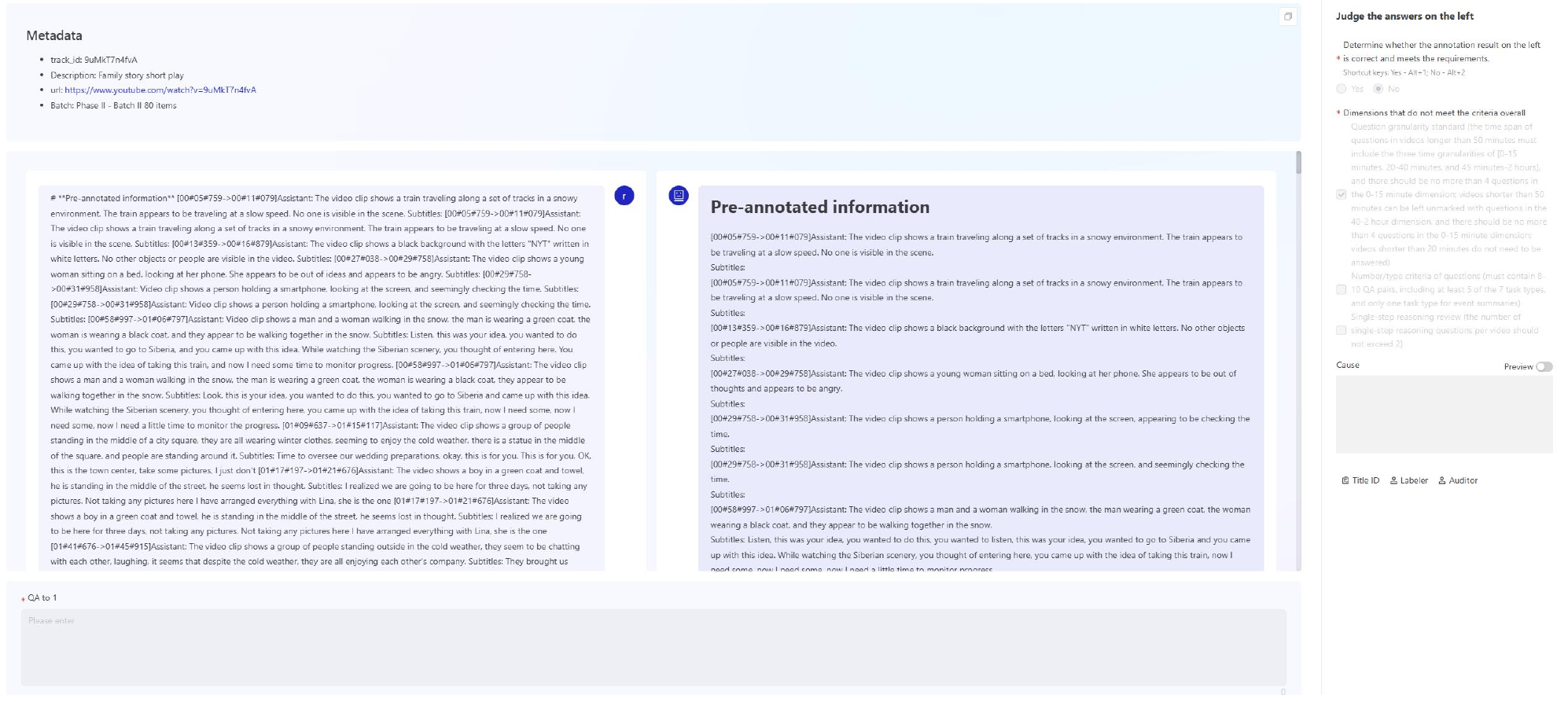}
   \caption{Interface of video-level evaluation.}
    \label{fig:validation_interface1}
\end{figure}
\begin{figure}[t]
  \centering
   \includegraphics[width=0.95\linewidth]{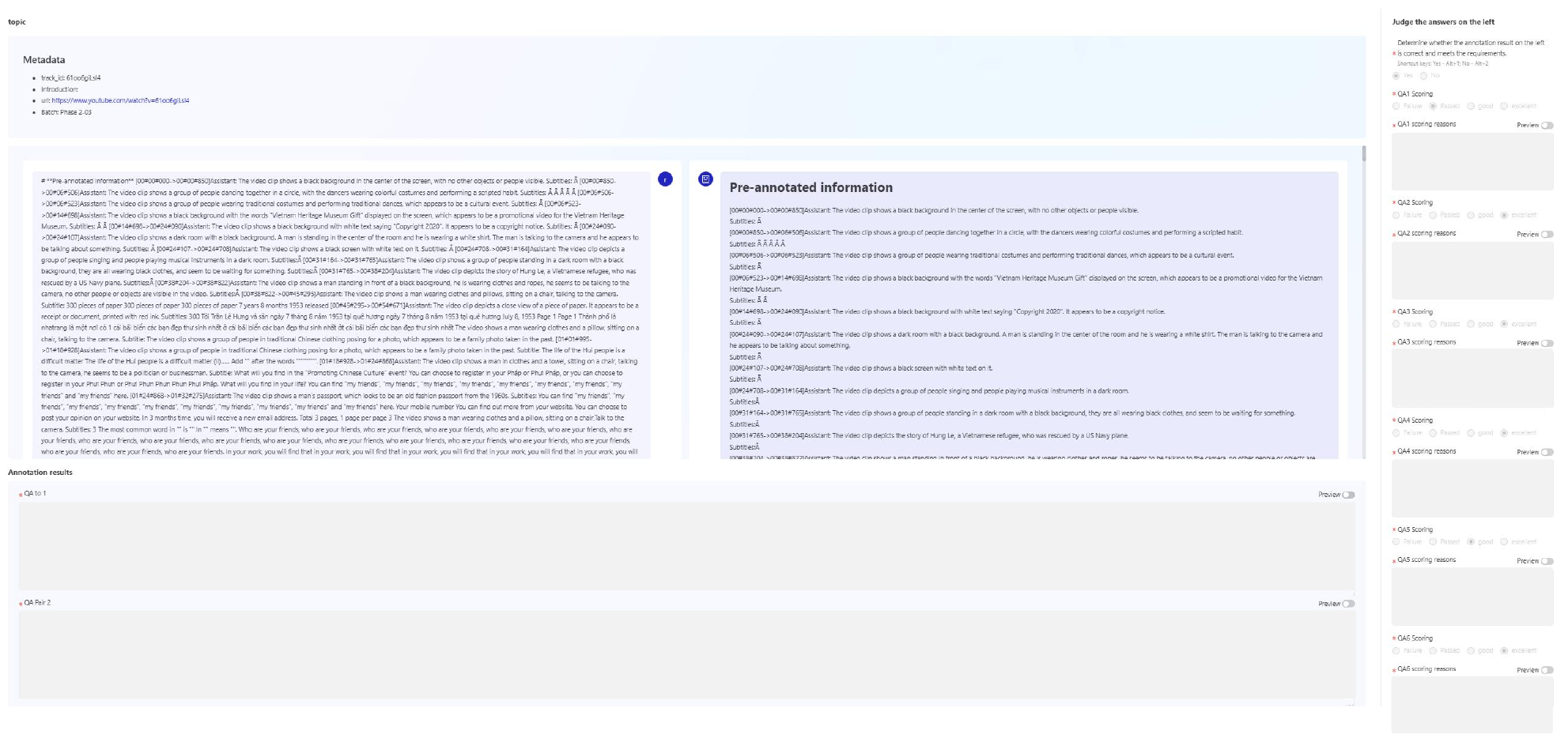}
   \caption{Interface of question-level evaluation.}
    \label{fig:validation_interface2}
\end{figure}

\subsection{Annotation Interface}

The annotation interface is illustrated in Figure~\ref{fig:annotation_interface}

\subsection{Details of Video Summary Generation}

The process of generating video summary involves two main steps. First, for each video segment, we merge visual descriptions with subtitles in chronological order while eliminating duplicate subtitles. If the subtitle content will be removed if it is overlapped with the former subtitle. This pattern continues until the final scene,

The second step entails generating video-level summary based on segment-level summaries. We set a 10-minute temporal interval to generate multiple segment-level summaries by integrating multimodal descriptions in each scene. During the integration, the information from previous segments, including both visual descriptions and subtitles, serves as the narrative background for each subsequent segment. The first segment-level summary becomes the narrative background for generating the second summary, which does not reiterate events from the first segment. This process continues iteratively until the final segment-level summary, resulting in a comprehensive summary of the video across all clips. 

\begin{figure}[t]
  \centering
   \includegraphics[width=0.95\linewidth]{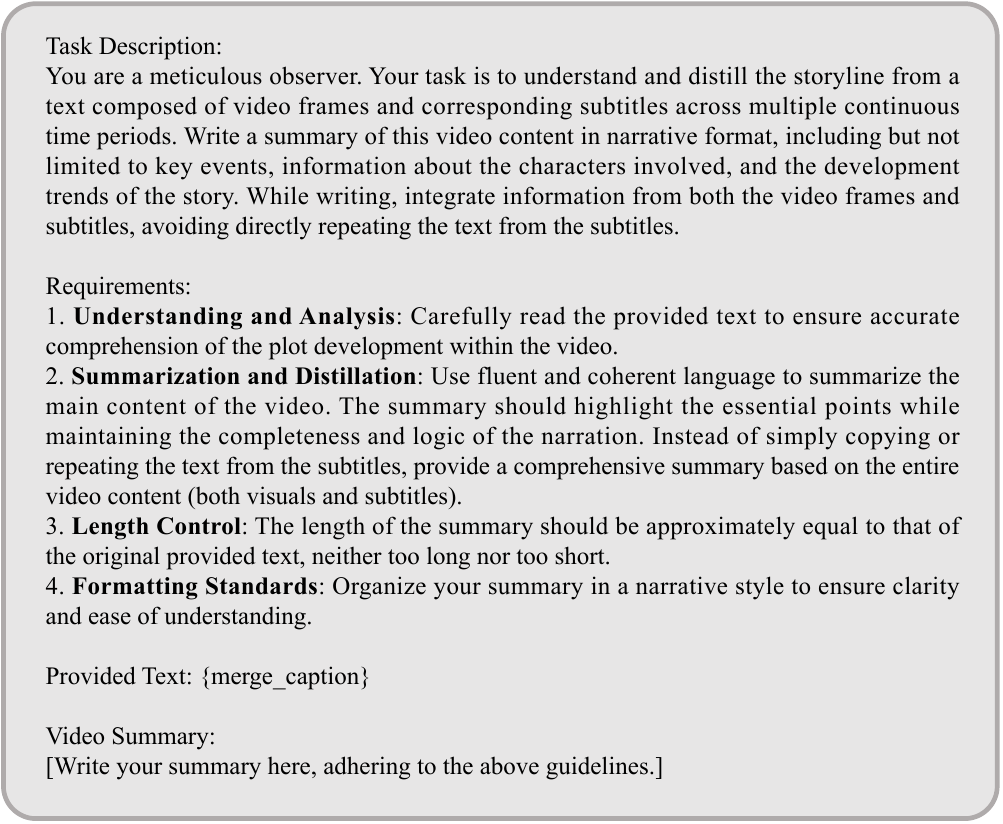}
   \caption{Prompts of generating the first video summary.}
    \label{fig:first_summary}
\end{figure}

\begin{figure}[t]
  \centering
   \includegraphics[width=0.95\linewidth]{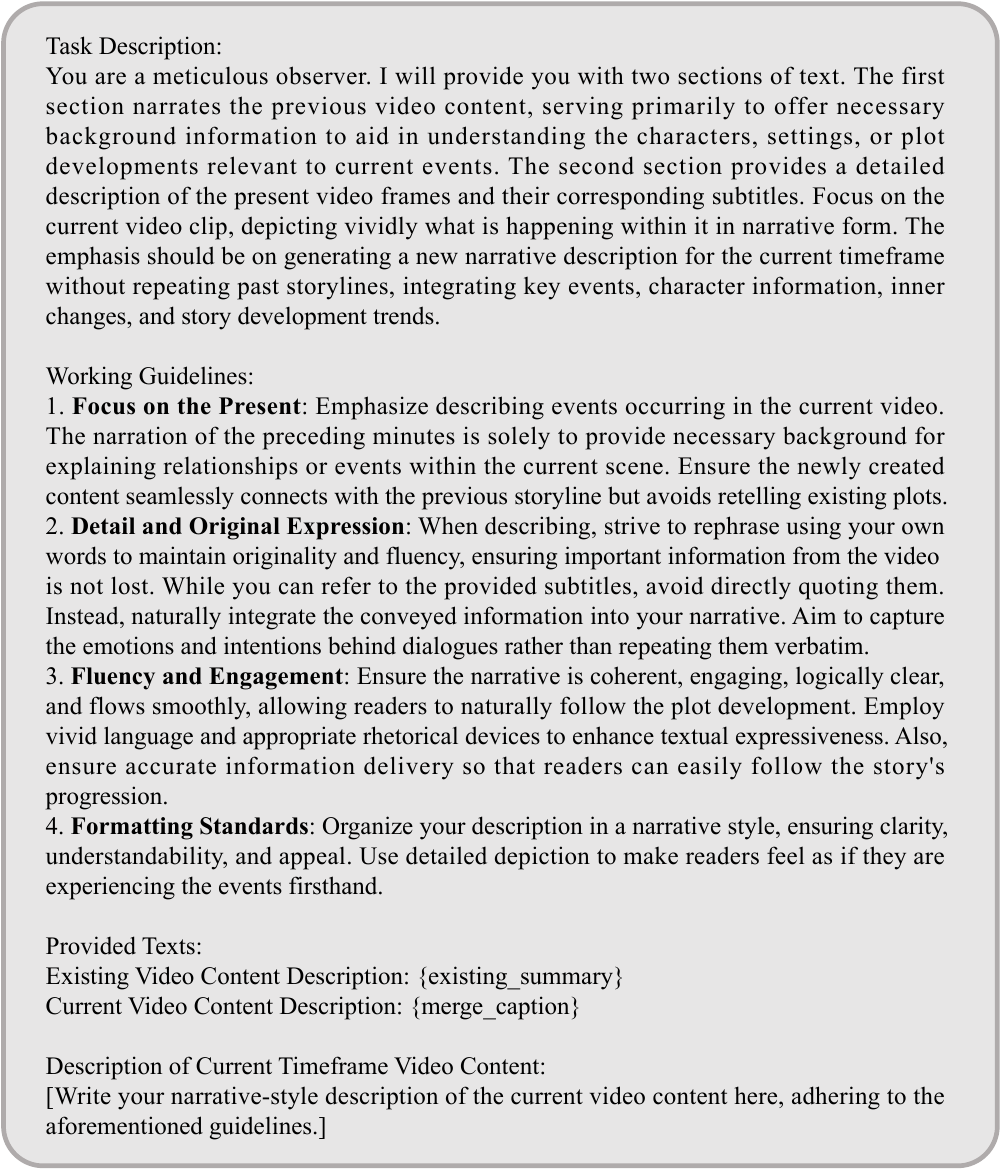}
   \caption{Prompts of generating other video summaries.}
    \label{fig:other_summary}
\end{figure}

We illustrate the prompt of generating the first video summary in Figure~\ref{fig:first_summary}, and the prompt of generating the following video summary in Figure~\ref{fig:other_summary}. We adopt Qwen-2.5-72B-Instruct~\cite{qwen2.5} to generate video summary.

\subsection{Prompts GPT pseudo annotations}
\begin{figure}[t]
  \centering
   \includegraphics[width=0.95\linewidth]{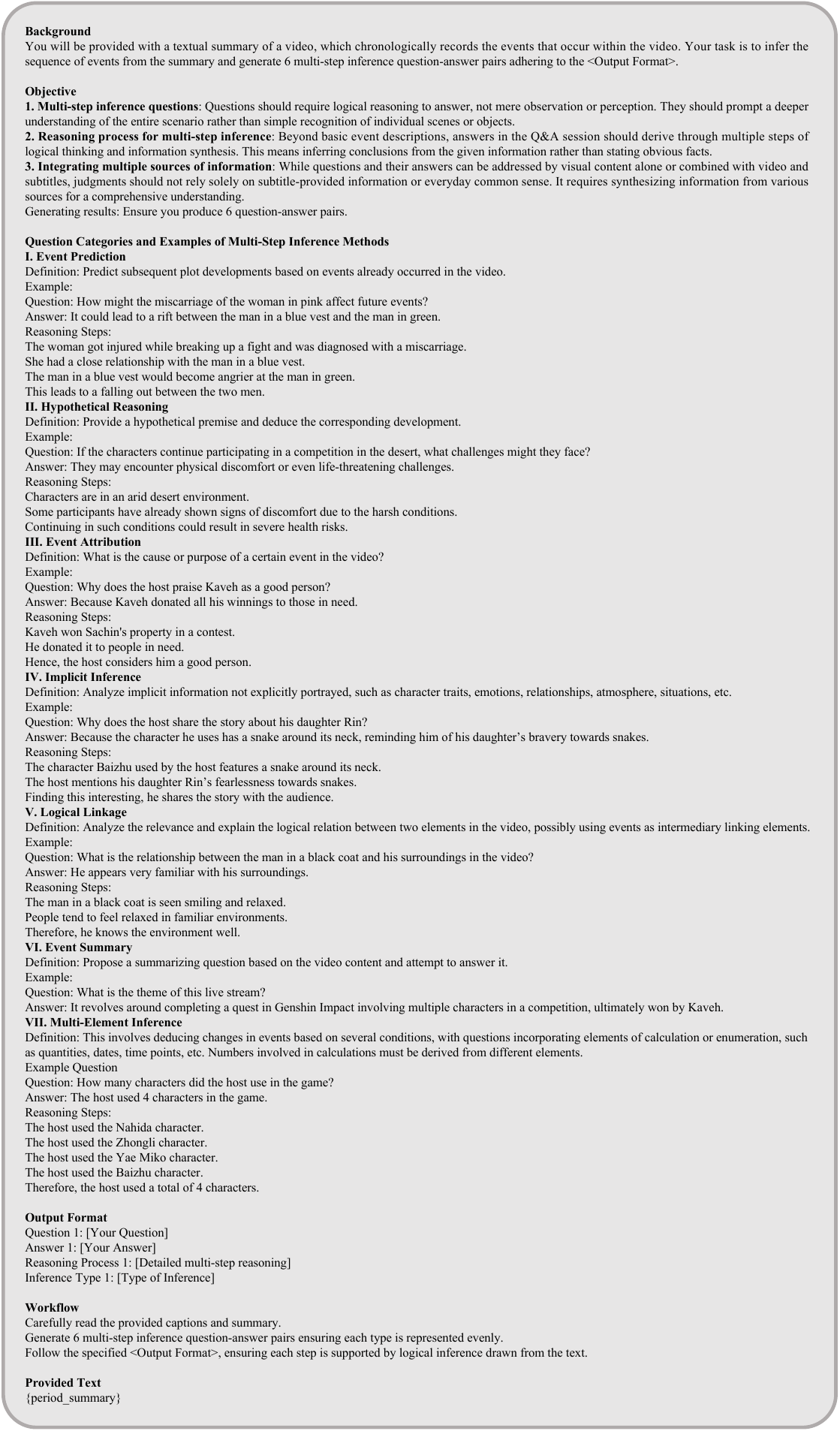}
   \caption{Prompts of generating GPT pre-annotation.}
    \label{fig:prompts_prelabel}
\end{figure}

We provide the prompts for generating GPT pseudo annotations in Figure~\ref{fig:prompts_prelabel}. During the generation process, we put the prompt with the video summary into GPT-4o~\cite{gpt4o}.

\section{Human Validation Details}
\label{appen:qa}

\subsection{Validation Guideline}

\subsubsection{Confidentiality Statement}
The content of this document involves trade secrets or other secrets, and is only for internal members and authorized units or individuals to review, please receive this document to keep it in a safe place, without the consent of any third party shall not be disclosed to this document.

\subsubsection{Task Overview}

Given a video, translated subtitles, 6 pre-generated pseudo annotations, and 8-10 manually generated multi-step reasoning annotations specific to the video content, the quality assessor needs:
\begin{enumerate}
    \item Give a unqualified/qualified binary video-level rating for the question distributions and provide reasons for the unqualified annotations.
    \item Give a four-choice question-level rating of unqualified/normal/good/excellent for each QA pair and provide the rationale.
\end{enumerate}

\noindent\textbf{Validation Requirements.} The quality assessors need to follow the following criteria to determine whether the given set of annotations is qualified or not. The quality assessors need to:
\begin{itemize}
    \item Determine whether the 8-10 QA pairs of the entire video meet the video-level QA standards. If not, directly choose to fail under the first question and provide the reason for failure without proceeding to the second step of QA.
    \item Determine whether each question under the entire video meets the following question-level quality control standards. If all meet, it will be judged as correct labeling; if one or more do not meet, it will be judged as unqualified labeling, and the reasons for failure must be provided in the prescribed format.
\end{itemize}

If the annotation is judged to be correct, proceed to question-level validation. Otherwise, check the radio box: failed, and in the following text box, provide the specific number of non-compliance with the quality control standards and reasons. It is necessary to write out all non-compliant QC standard entries to facilitate subsequent changes by the labeler.

\noindent\textbf{Video-Level Quality Control Standard.}
\begin{enumerate}[label=\alph*.]
    \item \textbf{Time Granularity Audit:} Each video question time span must contain [0-15min, 15-40min, 40min-2h] the three time granularities of the question, and 0-15min questions should not exceed 4.
    \begin{itemize}
        \item Videos shorter than 20 minutes can be marked as "unanswerable" and will be reviewed without moderation.
        \item Videos shorter than 50 minutes do not need to mark the 40min-2h dimension problem; the review only needs to determine whether the other two time dimension conditions are met.
    \end{itemize}
    \item \textbf{Number/Type of Questions Reviewed:} Each video must contain 8-10 QA pairs, including at least 5 of the 7 types of tasks. There can only be one task type for information synopsis.
    \item \textbf{Single-Step Reasoning Review:} No more than 2 single-step reasoning questions per video.
\end{enumerate}

\subsection{Configurations of Evaluated Models}

We provide the detailed configuration of evaluated VLM, including the model version and input visual frames, which is illustrated in Figure~\ref{tab:multimodal_model_configuration}.

\begin{table*}[h]
\centering
\footnotesize
\resizebox{\textwidth}{!}{%
\begin{tabular}{llllcrcc}
\toprule
\textbf{Organization} & \textbf{Model} & \textbf{Release} & \textbf{Version} & \textbf{\begin{tabular}[c]{@{}c@{}}Support\\Video?\end{tabular}} & \textbf{\begin{tabular}[c]{@{}c@{}}Input\\Frames\end{tabular}} & \textbf{\begin{tabular}[c]{@{}c@{}}\# multi-step reasoning\\Pipeline\end{tabular}} & \\
\midrule
\multicolumn{8}{l}{\emph{\textbf{Proprietary Models}}} 
\\
 \midrule
\multirow{1}{*}{OpenAI} & GPT-4o~\cite{gpt4o} &  2024-8  &  \texttt{gpt-4o-2024-08-06} & \ding{55} & 64 & API \\
\noalign{\vskip 0.5ex}\hdashline\noalign{\vskip 0.5ex}

\multirow{1}{*}{Google} & Gemini 2.0 Pro~\cite{gemini} & 2025-2 & \texttt{Gemini 2.0 Pro} & \ding{51} & 0.5fps & API \\
\noalign{\vskip 0.5ex}\hdashline\noalign{\vskip 0.5ex}

\multirow{1}{*}{Anthropic} & Claude-3.7-Sonnet~\cite{claude} & 2025-2 & \texttt{Claude-3.7-Sonnet-No-extended-thinking} & \ding{55} & 20 & API\\
\midrule
\multicolumn{8}{l}{\emph{\textbf{Open-source Multimodal Foundation Models}}} \\
 \midrule
 
\multirow{3}{*}{Alibaba} & Qwen2-VL-7B~\cite{qwen2vl} & 2024-8 & \texttt{Qwen2-VL-7B-Instruct} &\ding{51} & 128 & \multirow{3}{*}{HF}\\
& Qwen2.5-VL-7B~\cite{qwen2.5vl} & 2025-1 & \texttt{Qwen2.4-VL-7B-Instruct} & \ding{51} & 128 & \\
& Qwen2.5-VL-72B~\cite{qwen2.5vl} & 2025-1 & \texttt{Qwen2.5-VL-72B-Instruct} & \ding{51} & 128 & \\
\noalign{\vskip 0.5ex}\hdashline\noalign{\vskip 0.5ex}

\multirow{2}{*}{Shanghai AI Lab} & InternVL2.5-78B~\cite{internvl2.5} & 2024-11 & \texttt{InternVL2.5-78B} & \ding{55} & 64 & \multirow{2}{*}{HF}\\
& InternVL2.5-8B~\cite{internvl2.5} & 2024-11 & \texttt{InternVL2.5-8B} & \ding{55} & 64 & \\
\noalign{\vskip 0.5ex}\hdashline\noalign{\vskip 0.5ex}

\multirow{2}{*}{OpenGVLab} & InternVideo2.5-8B~\cite{internvideo2.5} & 2025-2 & \texttt{InternVideo2.5-Chat-8B} & \ding{51} & 64 & \multirow{2}{*}{HF} \\
& VideoChat-Flash~\cite{videochatflash} & 2025-1 & \texttt{VideoChat-Flash-7B@224} & \ding{51} & 512 & \\
\noalign{\vskip 0.5ex}\hdashline\noalign{\vskip 0.5ex}

\multirow{1}{*}{Microsoft} & Phi-3.5-Vision~\cite{phi} & 2024-7 & \texttt{Phi-3.5-vision-instruct} & \ding{55} & 8 & HF \\
\noalign{\vskip 0.5ex}\hdashline\noalign{\vskip 0.5ex}

\multirow{1}{*}{Rhymes} & Aria~\cite{aria} & 2024-11 & \texttt{Aria-Chat} & \ding{55} & 8 & HF \\
\noalign{\vskip 0.5ex}\hdashline\noalign{\vskip 0.5ex}

\multirow{1}{*}{H2O} & H2OVL Mississippi-2B~\cite{h2ovl} & 2024-10 & \texttt{h2ovl-mississippi-2b} & \ding{55} & 4 & HF \\
\noalign{\vskip 0.5ex}\hdashline\noalign{\vskip 0.5ex}

\multirow{1}{*}{DeepSeek} & DeepSeek-VL2~\cite{deepseekvl2} & 2024-12 & \texttt{deepseek-vl2 } & \ding{55} & 4 & HF \\
\noalign{\vskip 0.5ex}\hdashline\noalign{\vskip 0.5ex}

\multirow{2}{*}{lmms-lab} & LongVA-7B~\cite{longva} & 2024-6 & \texttt{longVA-7B} & \ding{51} & 128 & \multirow{2}{*}{HF} \\
& LongVA-7B-DPO~\cite{longva} & 2024-6 & \texttt{LongVA-7B-DPO} & \ding{51} & 128  & \\
\noalign{\vskip 0.5ex}\hdashline\noalign{\vskip 0.5ex}

\multirow{1}{*}{Tencent} & ARC-Hunyuan-Video-7B~\cite{archunyuan} & 2025-7 & \texttt{ARC-Hunyuan-Video-7B} & \ding{51} & 150 & HF\\
\noalign{\vskip 0.5ex}\hdashline\noalign{\vskip 0.5ex}

\multirow{1}{*}{Kwai} & Keye-VL-8B-Preview~\cite{keyevl} & 2025-6 & \texttt{Keye-VL-8B-Preview} & \ding{51} & 128 & HF\\
\noalign{\vskip 0.5ex}\hdashline\noalign{\vskip 0.5ex}

\multirow{1}{*}{Moonshot} & Kimi-VL-A3B-Thinking~\cite{team2025kimi} & 2025-7 & \texttt{Kimi-VL-A3B-Thinking-2506} & \ding{51} & 256 & vLLM\\
\noalign{\vskip 0.5ex}\hdashline\noalign{\vskip 0.5ex}

\multirow{1}{*}{Xiaomi} & MiMo-VL-7B-RL~\cite{mimovl} & 2025-5 & \texttt{MiMo-VL-7B-RL} & \ding{51} & 512 & vLLM\\

\bottomrule
\end{tabular}
}

\caption{
Details of evaluated multimodal foundation models. The ''Input Frames'' column represents the default number of input frames, chosen from {4, 8, 16, 32, 64, 128, 256, 512}, based on the maximum value that does not exceed the model's context window and the constraints of GPU memory. ``HF'' means ``Hugging Face''.
}
\label{tab:multimodal_model_configuration}
\end{table*}

\noindent\textbf{Problem-Level Quality Control Standard.}
\begin{enumerate}[label=\alph*.]
    \item \textbf{Question Type Review:} The type of question must be a reasoning question, not a perception question.
    \begin{itemize}
        \item \textbf{Perceptual Questions:} Questions that can be derived directly from the picture.
        \item \textbf{Reasoning Problems:} Problems that require reasoning in conjunction with perceptual results.
    \end{itemize}
    \item \textbf{Question Modal Audit:} The answer to a question must be inferred from the video content or video + subtitle content, not from the subtitles alone.
    \item \textbf{Annotation Format Review:} The annotation must contain the following five items: the corresponding start time of the question, the question, the answer, the multi-step reasoning, and the timestamp of each step (for question types such as hypothetical reasoning and event prediction, the timestamp of the step can be omitted if these virtual events do not occur in the video), the question type. Answers must conform to the given template.
    \item \textbf{Question Correctness Audit:} According to the labeled video start time, the answer to the question can be obtained through the video content, and the reasons and problem categories given by the quality inspector are reasonable.
    \item \textbf{Reasoning Step Audit:} Except for the 0-15min time granularity and information synopsis task, all questions must be multi-step reasoning problems. If you find that the contents of the reasoning steps are too similar to other steps, you can merge the reasoning steps, and the final reasoning step numbers will be calculated according to the merged reasoning steps. The number of reasoning steps is equal to the number of reasoning process bars - 1.
    \item \textbf{Timestamp Redundancy Audit:}
    \begin{itemize}
        \item For issue timestamp markers, time redundancy should not exceed 5 minutes.
        \item For reasoning step timestamp markers, time redundancy must not exceed 1 minute.
        \item Example: If for a manually generated question, the latest time node to start watching is $t_1$, and the earliest time node that can end watching is $t_2$, then the question's timestamp is labeled as [$t_1$->$t_2$]. If the manually generated question's timestamp is [$t_3$->$t_4$], the question is labeled as failing when $t_3+5<t_1$ or $t_4-5>t_2$.
    \end{itemize}
\end{enumerate}

\noindent\textbf{Step 2: Labeling Quality Scores.} 
\noindent\textbf{Quality Control Standards:}
\begin{enumerate}[label=\alph*.]
    \item \textbf{Reasoning Step QC:} 5 or more steps of reasoning meet the excellent rating, 3-4 steps of reasoning are labeled as good, and 1-2 steps of reasoning are labeled as normal. (Note: Questions that do not meet the QC criteria for the reasoning step review should be directly labeled as failing and will not participate in scoring.)
    \item \textbf{Question Difficulty QC:} Subjective judgment of the difficulty of questions posed by annotators. Questions that are easy to answer qualify for a normal grade, those that are more difficult to answer qualify for a good grade, and those that are very difficult qualify for an excellent grade.
    \item \textbf{Reasoning Process Uniqueness Judgment:} If the reasoning process conforms to uniqueness, i.e., the answer given by the user cannot be reached by other reasoning steps, then it conforms to the excellent grade. A normal grade is met if it is clear that some other reasonable reasoning process could have led to the answer given by the user; a good grade is met if it is not clear whether the answer is unique or not.
    \item \textbf{Reasoning Process Detail QC:} If each reasoning step is a concise, complete sentence, it meets the excellent rating. If the reasoning is redundant and complete in one or more sentences per step, or if the reasoning is concise but less than complete in one sentence per step, it meets the good rating. If each step of reasoning is neither concise nor complete, it meets the normal grade.
\end{enumerate}

Following the above four quality control standards for grade determination, the final score is calculated: 1 point for excellent, 0 points for good, -1 point for normal. If the total score is greater than 0 points, it will be labeled as excellent; if the total score is equal to 0 points, it will be labeled as good; and if the total score is less than 0 points, it will be labeled as normal. 

\subsubsection{Quality Assessment}
We randomly sample  human validation results with a portion of 5\%-10\% for quality assessment following these criteria:
\begin{enumerate}
    \item Assessing correctness to determine compliance with quality control standards.
    \item Assessing quality scores for compliance with validation standards.
    \item Whether each manual annotation is scored, with or without reasons for vacant scoring.
\end{enumerate}

\subsection{Validation Interface}

The video-level annotation interface is illustrated in Figure~\ref{fig:validation_interface1}, and we show the question-level annotation interface in Figure~\ref{fig:validation_interface2}.

\section{Evaluation Setup}

\subsection{Prompts for Model Inference}

\begin{figure}[htbp]
  \centering
  \begin{subfigure}[b]{0.95\linewidth}
    \includegraphics[width=\linewidth]{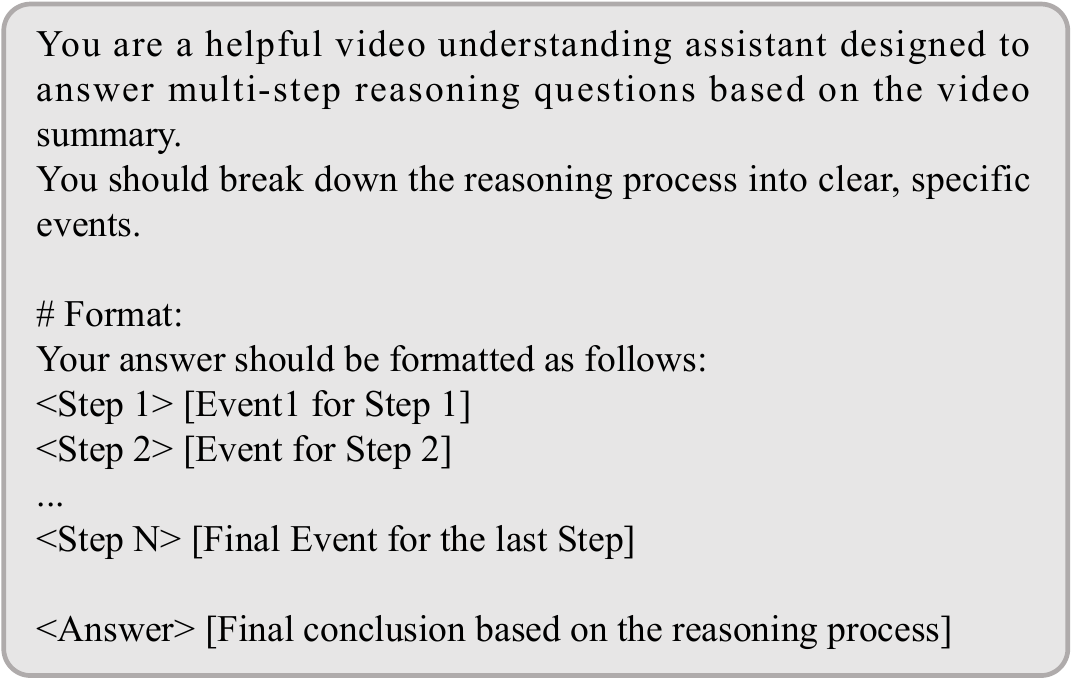}
    \caption{System prompt for LLM}
    \label{fig:single_mcq-1st_round-LLM-COT_SYSTEM_PROMPT}
  \end{subfigure}  
  \begin{subfigure}[b]{0.95\linewidth}
    \includegraphics[width=\linewidth]{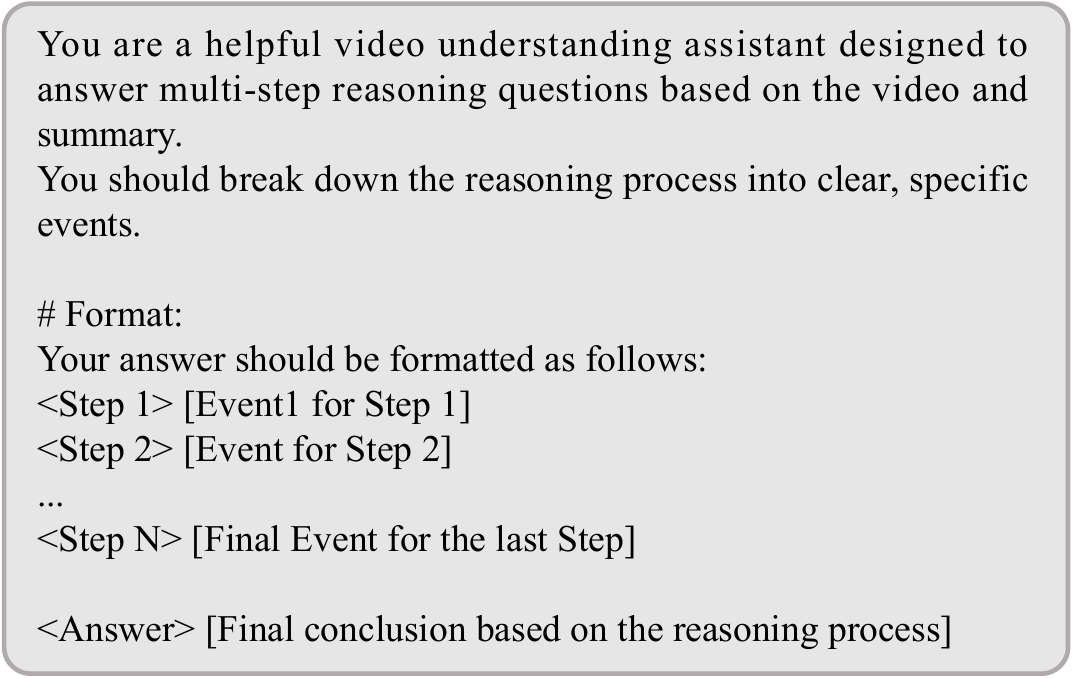}
    \caption{System prompt for VLM}
    \label{fig:single_mcq-1st_round-MLLM-COT_SYSTEM_PROMPT}
  \end{subfigure}  
  \vspace{\floatsep} %
  \begin{subfigure}[b]{0.95\linewidth}
    \includegraphics[width=\linewidth]{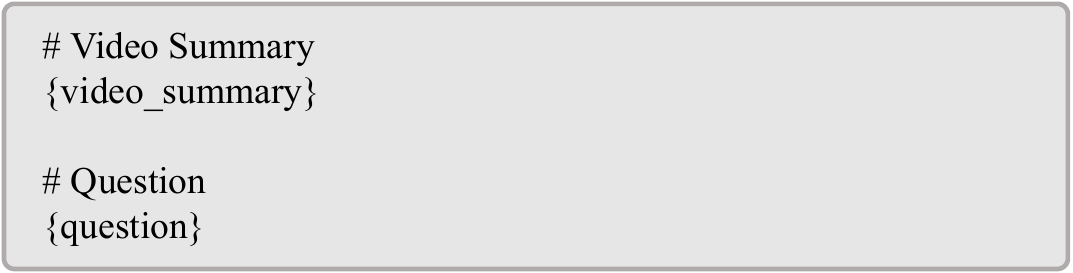}
    \caption{User prompt template}
    \label{fig:single_mcq-1st_round-COT_HUMAN_PROMPT_TEMPLATE}
  \end{subfigure} 
  \caption{Prompt for LLM Inference} 
  \label{fig:single_mcq_1st_round}
\end{figure}

We here provide prompts for model inference in Figure~\ref{fig:single_mcq_1st_round} for LLM and VLM. 

\subsection{Prompts for Open-ended Evaluation}

\begin{figure}[htbp]
  \centering
   \includegraphics[width=0.95\linewidth]{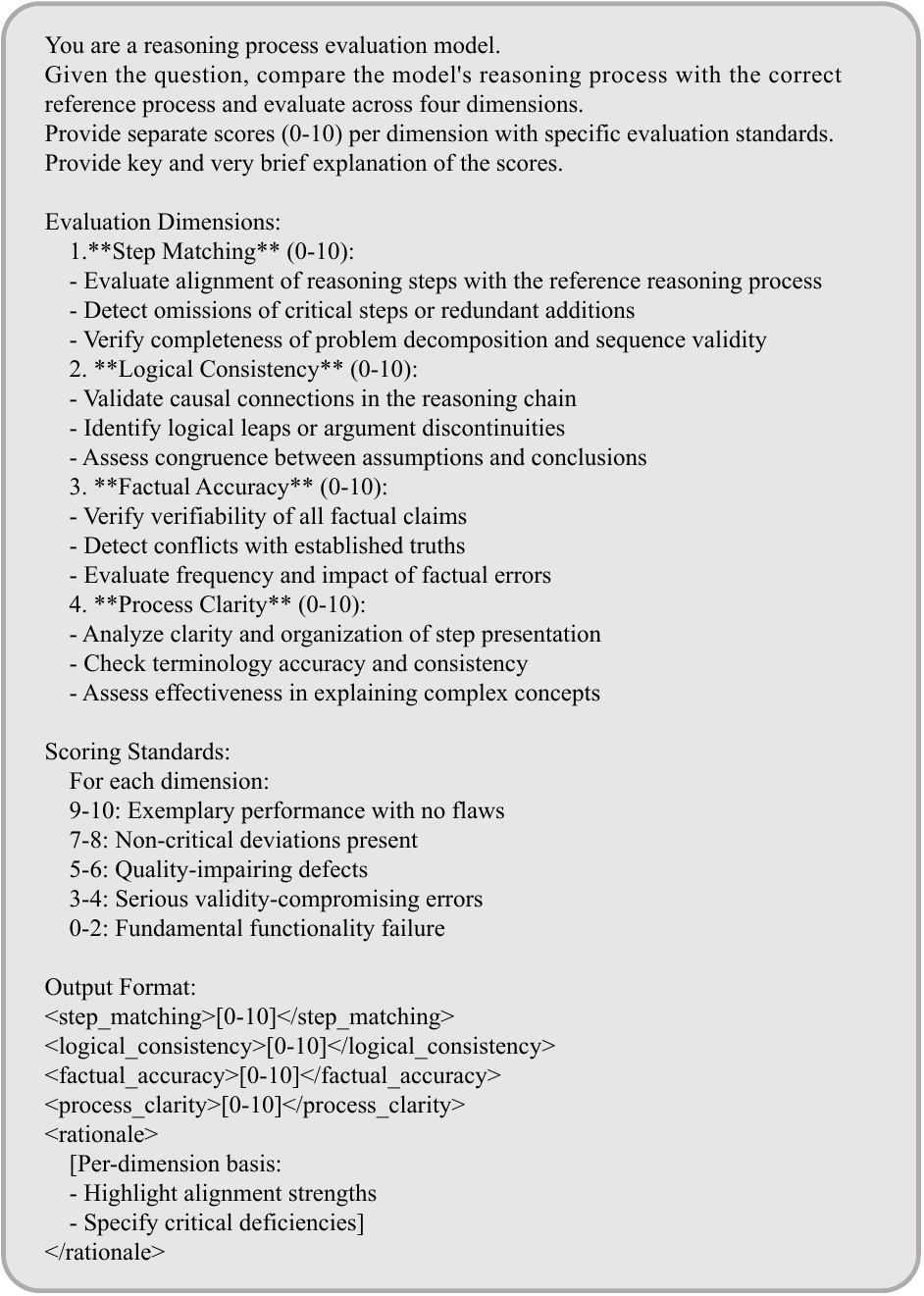}
   \caption{System prompt for a question having a unique answer}
    \label{fig:UNIQUE_ANSWER_EVAL_SYSTEM_PROMPT}
\end{figure}

Prompts for open-ended evaluation are illustrated in Figure~\ref{fig:UNIQUE_ANSWER_EVAL_SYSTEM_PROMPT}.

\section{Ethical Discussion}

\subsection{Data Safety Discussion}
\label{appen:safety}
VRBench is constructed from a large number of long narrative videos that may include harmful visual content and speech transcripts. To avoid potential ethical issues, we employ a multimodal unsafe detection strategy. First, we extract frames at 1 fps and use an image safety detection API to filter videos that include pornography, horror, politics, violence, and vulgarity content. Then we use a multimodal language safety detection API to filter all vanilla speech transcripts and video captions. We use the same API tags as image unsafe detection to filter textual data. During manual filtering, we also explicitly request the human annotators to discard videos deemed to contain terror, pornographic, or political content for data safety assurance. 

\subsection{Dataset Bias}
VRBench includes videos in 8 languages that exclude English and Chinese, the two most widely used languages around the world. Therefore, VRBench may contain potential cultural bias for countries and regions where these languages are used. During data curation, we strive to ensure that all races, skin colors, and genders are included in the data collection scope, which demonstrates that we have not exhibited but actively avoided any subjective bias regarding these characteristics.

\subsection{Potential Social Impact}
We argue that VRBench does not contain elements that could pose a potential societal negative impact. We have implemented the aforementioned measures in Section~\ref{appen:safety} to prevent the inclusion of harmful information, and we will make every effort to revise the dataset if any content is found to produce unforeseen negative impacts. On the contrary, VRBench might draw the AI research's focus more to the area of non-Chinese and non-English communities, which we believe is currently being overlooked by published datasets. We consider this to be a potential socially beneficial aspect.

\subsection{LLM Usage Claim}
We use LLMs for proofreading the writing of the paper. We have verified that all content reflects the authors' original intent and does not contain factual errors or hallucinated information from LLMs.

\end{document}